# Fast single image super-resolution based on sigmoid transformation

Longguang Wang, Zaiping Lin, Jinyan Gao, Xinpu Deng, Wei An[1]

***Abstract*—Single image super-resolution aims to generate a high-resolution image from a single low-resolution image, which is of great significance in extensive applications. As an ill-posed problem, numerous methods have been proposed to reconstruct the missing image details based on exemplars or priors. In this paper, we propose a fast and simple single image super-resolution strategy utilizing patch-wise sigmoid transformation as an imposed sharpening regularization term in the reconstruction, which realizes amazing reconstruction performance. Extensive experiments compared with other state-of-the-art approaches demonstrate the superior effectiveness and efficiency of the proposed algorithm.**

## I. Introduction

SUPER-resolution (SR) has been widely applied in video surveillance [1][2][8], remote sensing [3][4][5][35], medical imaging [6] [7] and many other fields, enhances the image resolution and provides pleasing image details, which is significant for subsequent image processing. Aiming to upscale the image resolution based on single image [12], single image super-resolution (SISR) requires generating severalfold data from limited input data to approach natural images, which is badly ill-posed and often suffers from annoying blurring and artifacts.

To alleviate the ill-posed problem of SISR, many algorithms have been proposed to exploit additional information to learn how natural images are, which improves the reconstruction performance efficiently. Considering the difference between natural image information sources, the algorithms can be roughly subdivided into dictionary-based methods [9][17][18][33][34], self-exemplar-based methods [15][20][32] and prior-based methods [13][14][16]. Dictionary-based methods refer to external high resolution (HR) and corresponding low resolution (LR) image pairs as exemplars to hallucinate the missing details, which requires large scale databases to cover possible relationship between LR images and HR images. As external-exemplars-based methods are time-consuming in training procedure, self-exemplar-based methods assume redundancy of patches within the single image and utilize recurred local similar patches as exemplars to exploit underlying image details. Unlike dictionary-based and exemplar-based, prior-based methods utilize priors as constraints to alleviate the ill-posedness, perform robustly and efficiently with no relying on exemplars.

Interpolation methods [22][23][26] as the simplest prior-based methods, utilize analytical interpolation formulae like bicubic scheme to predict new pixels based on local pixels, however the spline functions may not match natural images with strong discontinuities and lead to ringing artifacts along the edges. Smoothness prior [24][25] as another widely used category of image prior, regularizes the first or higher derivatives of the reconstructed image, suppressing the additional noises effectively but leading to transparent blurring. Recently sparsity deduced priors [14][21][27][28][31] have been extensively investigated which assume local image patch can be sparsely represented by linear combination of over-complete dictionary, however the sparsity assuming loses texture and other image details. To alleviate the blurring effect of edges in SISR, edge-based priors [11][13][16] are introduced to SISR and realize outstanding sharpening and deblurring effect. Fattal [11] proposed an edge deduced prior based on statistics of edge features, imposed the local continuity measures in the upscaled image to match the statistics learned from HR and LR image pairs. Sun [13][16] proposed a novel gradient profile prior, utilized 1-D profiles of gradient magnitudes to describe the gradient structure with a parametric gradient profile model learned from a large scale of natural images.

Considering upscaling leads to little degradation in flat region and dominant deterioration is mainly concentrated in edge and texture regions performing as blurring effect, we are motivated to utilize sharpening operation to alleviate the degradation with no relying on external exemplars. In this paper we propose an adaptive patch-wise sigmoid transformation to realize appropriate sharpening of the upscaled image, then utilize the sharpened image as an imposed regularization in the reconstruction and realize amazingly effective deblurring and sharpness enhancement. Different from [13][16] implementing the sharpening process in gradient space, our patch-wise method utilizes sigmoid function to match the slope of local intensities directly and steepens the slope with sigmoid transformation, which is similar to the image enhancement in [29][30]. With this simple but effective process, our method produces state-of-the-art SISR results with superior processing efficiency, and the reconstructed images are sharp and distinct with rare artifacts.

The reminder of this paper is organized as follow: In **Section II** we first introduce the patch-wise sigmoid transformation. Then in **Section III** we present the SISR framework based on sigmoid transformation. Extensive experiments are conducted comparing with state-of-the-art algorithms in **Section IV**. Finally the conclusions are drawn in **Section V**.

L. Wang is with the College of Electronic Science, National University of Defense Technology, Changsha, China. (wanglongguang15@nudt.edu.cn)

Z. Lin, J. Gao, X. Deng and W. An are also with the College of Electronic Science, National University of Defense Technology, Changsha, China.

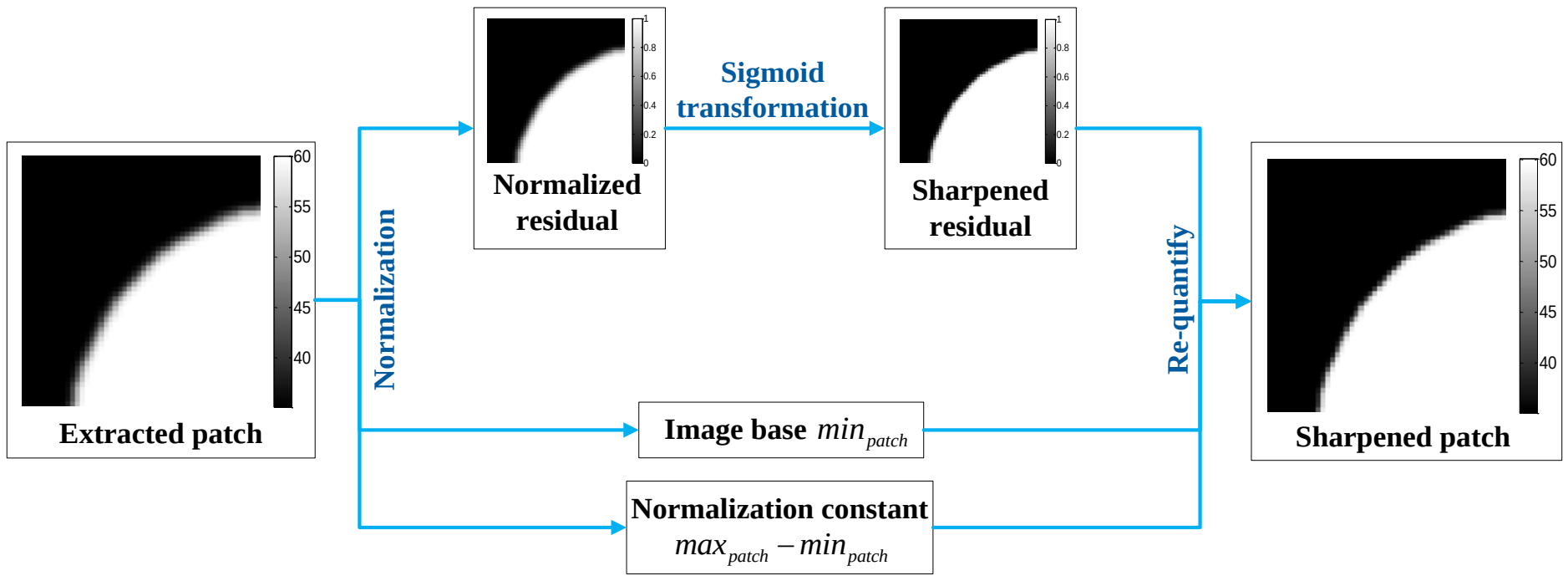

Fig.1 Overall sigmoid transformation.

## II. PATCH-WISE SIGMOID TRANSFORMATION

In this section, the proposed patch-wise sigmoid transformation is presented and analyzed. We first formulate the sigmoid transformation, then perform parametrical analysis and finally compare our sharpening operation to other edge enhancement methods.

### *A. Formulation*

In our scenario, the sharpening operation is implemented directly in image space instead of gradient space. Considering the continuity of natural images in local region, extracted local image patch can be regarded as a slope and the sharpening operation is equivalent to steepening the patch-wise slope intuitively. From the sketch of sigmoid function $f(x) = 1/(1+\exp(-a(x+b)))$ shown in **Fig. 2**, we can see that sigmoid functions with varying parameter pairs $a$ and $b$ perform as superior characterizations of slopes with ranged steepness and location, therefore we are motivated to choose it as fitting function to characterize patch-wise slope, and the slope-based steepening can be realized through simple parametrical transformation.

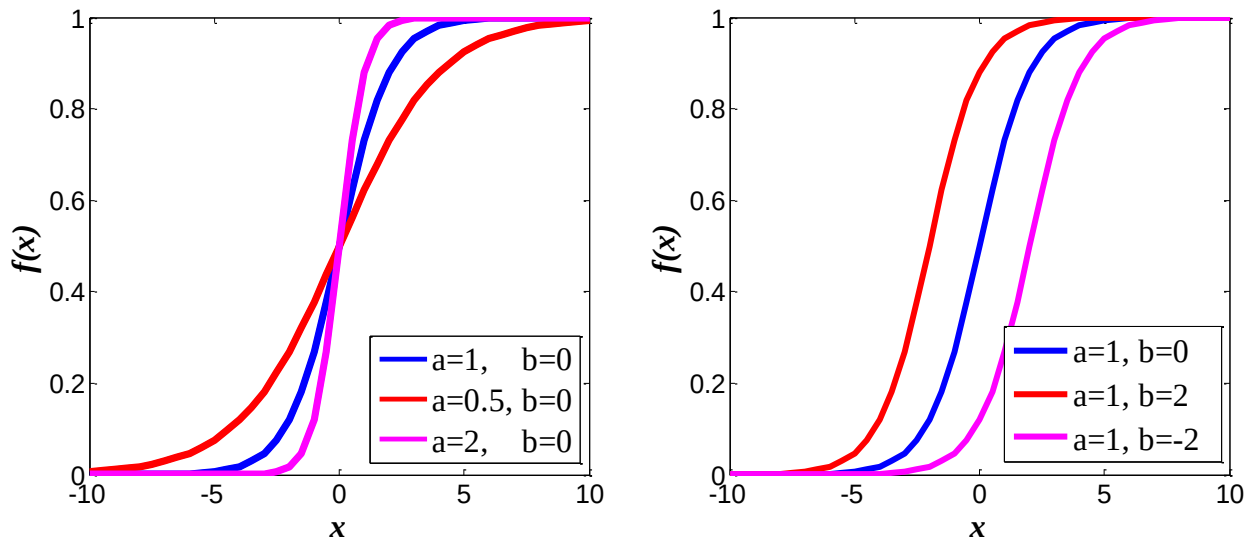

Fig.2 Sketch of sigmoid functions with ranged parameter pairs $a$ and $b$.

As shown in **Fig. 2**, the value of sigmoid function $f(x) = 1/(1+\exp(-a(x+b)))$ ranges from 0 to 1, therefore the intensities within the patch need to be normalized first for subsequent sigmoid transformation.

$$y_{i,i\in patch} = \frac{z_i - \min_{j\in patch}\{z_j\}}{\max_{j\in patch}\{z_j\} - \min_{j\in patch}\{z_j\}}, \quad (1)$$

where $z_i$ is the intensity of location $i$, $y_i$ represents the normalized value, and $\min_{j\in patch}\{z_j\}$,$\max_{j\in patch}\{z_j\}$ refer to the minimum and maximum of the patch respectively. To avoid $y_i$ equals to 0 or 1, we utilize a small constant $\varepsilon$ set to be 0.01 to fine-tune the maximum and minimum as

$$\begin{cases} \min_{j\in patch}\{z_j\} = \min_{j\in patch}\{z_j\} - \varepsilon \\ \max_{j\in patch}\{z_j\} = \max_{j\in patch}\{z_j\} + \varepsilon \end{cases}. \quad (2)$$

With normalization operation implemented, sigmoid function $f_0(x) = 1/(1+\exp(-a_0(x+b_0)))$ is introduced to fit the normalized residuals $y_i$ to derive corresponding $x_i$ values

$$y_i = f_0(x_i) \quad \Rightarrow \quad x_i = \frac{\ln(\frac{1}{y}-1)}{-a_0} - b_0, \quad (3)$$

and then we transform $x_i$ to sharpened residuals $y_i'$ with sharpened sigmoid function $f_1(x) = 1/(1+\exp(-a_1(x+b_1)))$

$$y_i' = f_1(x_i) = \frac{1}{1+\left(\frac{1}{y_i}-1\right)^{\frac{a_1}{a_0}} \cdot e^{a_1(b_0-b_1)}} \quad (4)$$

Note that all intensities within the image patch are fused together for operation in this process, leading to strong robustness to noises and undulation.

Finally we re-quantify the sharpened residuals $y_i'$ to derive sharpened intensities $z_i'$

$$z_i' = y_i' \times C_i + \min_{j\in patch}\{z_j\} \quad (5)$$

where $C_i = \max_{j\in patch}\{z_j\} - \min_{j\in patch}\{z_j\}$ represents the normalization constant.

As we can see from (4), the patch-wise sigmoid transformation can be regarded as a four-parameter problem, namely the sharpened image patch can be derived as parameters $\{a_0, a_1, b_0, b_1\}$ determined. Further observe (4), we can see it can be rewritten as

$$y_i' = \frac{1}{1+(\frac{1}{y_i}-1)^K \cdot e^B} = g(y_i; K, B)$$

$$\text{where} \begin{cases} K = \frac{a_1}{a_0} \\ B = a_1(b_0 - b_1) \end{cases} \quad (6)$$

Then the overall sigmoid transformation also degenerates into a two-parameter function as

$$z_i' = h(z_i; K, B) \quad (7)$$

where $K$ determines the steepness of the sigmoid function while $B$ determines the location, namely the non-symmetry of the sigmoid function. The overall process of sigmoid transformation is illustrated in **Fig. 1**.

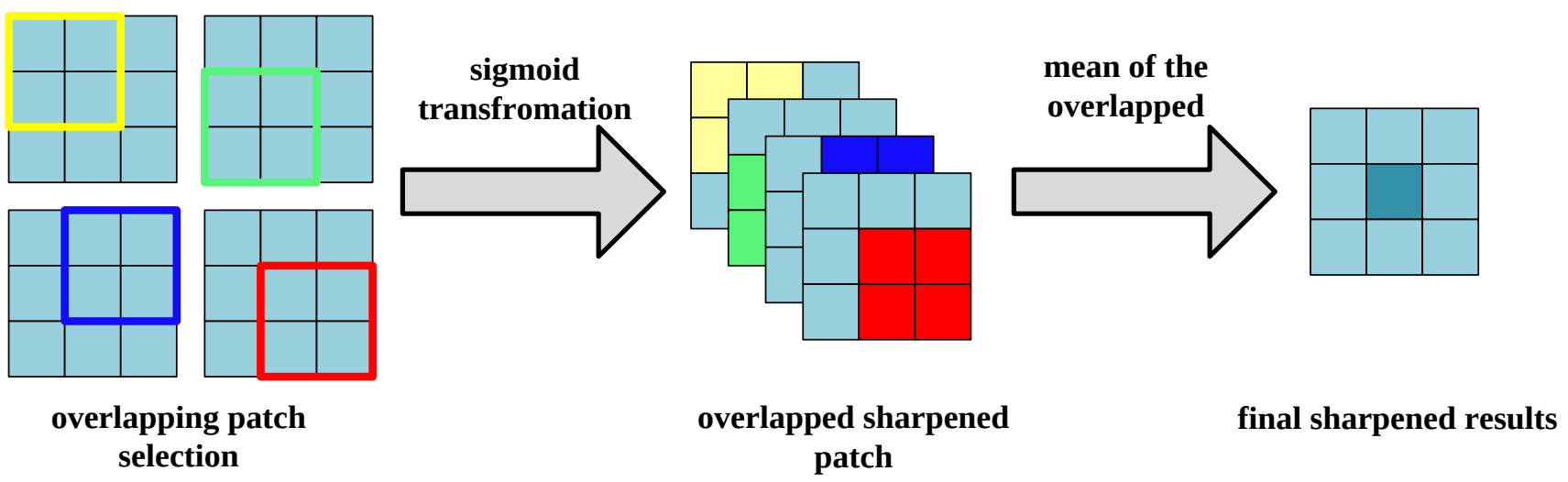

Fig.3 Overlapping strategy for patch-wise sigmoid transformation.

Considering patch-wise sigmoid transformation may generate block artifacts at the boundary between adjacent patches, we further utilize an overlapping strategy with Hanning window during implementation as shown in **Fig. 3**, and compute the mean value of overlapped pixels as the final sharpened results. As we can see, to implement the patch-wise sigmoid transformation with overlapping strategy, additional two parameters need to be determined, the size of image patch $l_{patch}$ and the stride $r_{patch}$.

*B. Parametrical analysis*

As analyzed above, totally four parameters consisting of $K, B, l_{patch}$ and $r_{patch}$ need to be determined for sigmoid transformation. In this part, the parameters are analyzed respectively and empirically determined, while more experimental results are exhibited in **Section V**.

● Sharpness $K$

Remembering parameter $K$ in (6) determines the steepness of the sigmoid function, it affects the sharpness and width of the edges directly. As we can see from left to right by row in **Fig. 4**, varying values of $K$ lead to ranged sharpening effects, the larger $K$ is, the sharper and narrower edges are. Moreover, it can be seen the sigmoid transformation performs as a blurring operation with smoothed and widened edges for $K < 1$, while performing as a sharpening operation with sharper and narrower edges for $K > 1$. Under the condition of $K = 1$, sigmoid transformation performs no sharpening or blurring with slope information well preserved in the results. Considering the requirement of image sharpening with computational cost, we empirically set $K = 2$ in this paper.

● Location $B$

Different with $K$, parameter $B$ determines the location of the sigmoid function as shown in **Fig. 2**, namely affects the extension of the edges. After downsampling operation, some edges may not locate at the center of LR pixel, which is often the case especially when upscaling factor $k$ is large. As shown in **Fig. 4**, we can see from top to bottom by column that the edges with varying values of $B$ are displaced along the vertical direction, and the sign of $B$ determines the direction. For $B < 0$ shown in the first row, the edges are displaced along the gradient direction, while displaced opposite to the gradient direction for $B > 0$ shown in the last row. Under the condition of $B = 0$, the edges are supposed to locate exactly at the center of pixel in LR image with no displacement during sharpening operation. Considering sub-pixel locations of edges in LR image vary between regions and are hard to estimate, especially when $k$ is large, therefore we empirically set $B = 0$ in this paper for simplicity.

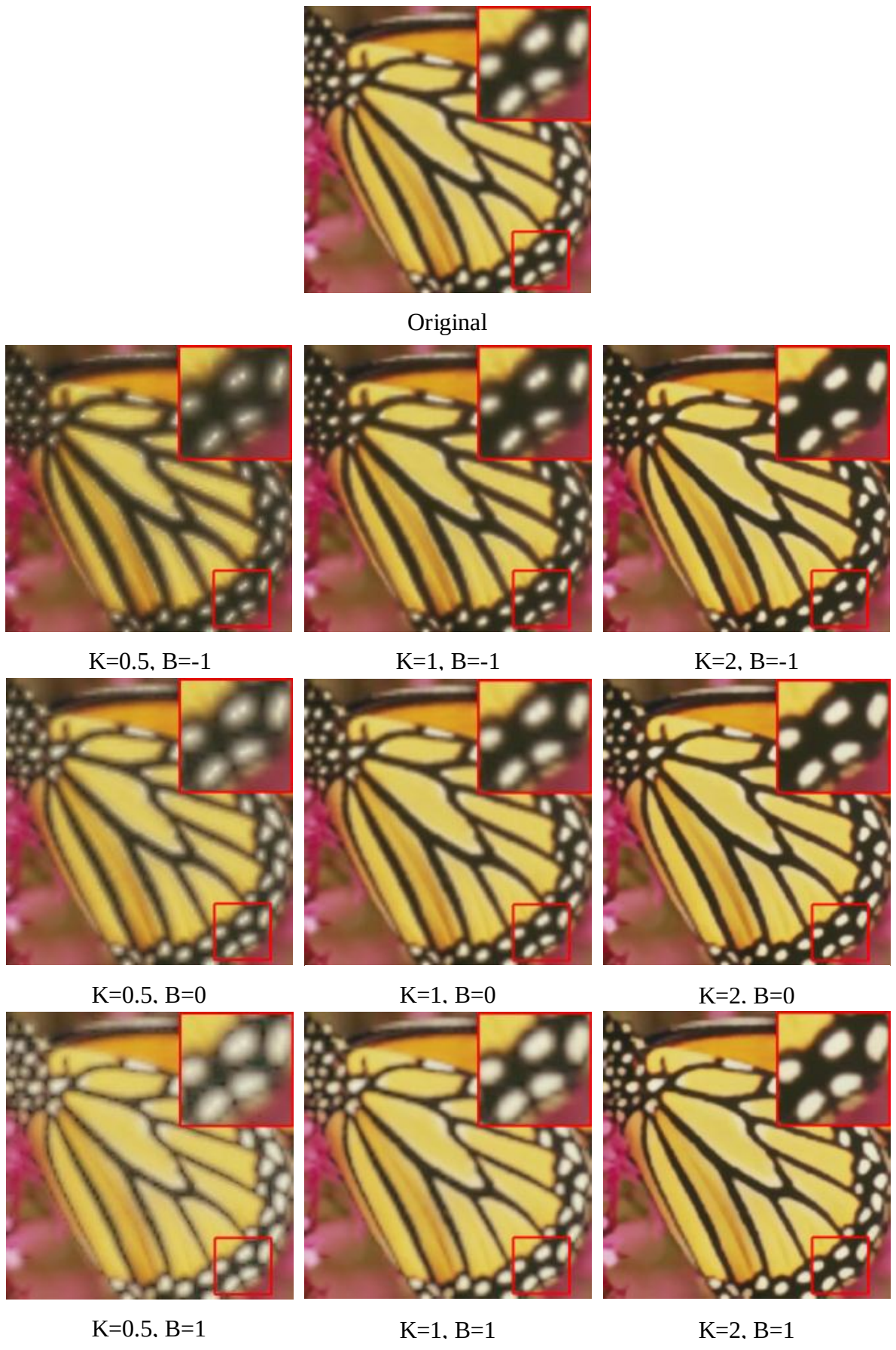

Fig.4 Effects with ranged values of $K$ and $B$. The results are derived from the blurred original image with different parameter settings.

● Patch size $l_{patch}$ and stride $r_{patch}$

As a patch-wise slope-based sharpening operation, sigmoid transformation requires the patch size $l_{patch}$ to cover complete slope information without disturbing information. Considering over-large patch size may introduce extra disturbing information and over small patch size may lose slope information, both leading to degradation of sharpening effect and generation of annoying artifacts, we empirically set $l_{patch} = 3$ in LR space to cover the neighboring information with little loss of slope information. To reduce the block artifacts at the boundary between patches, stride $r_{patch}$ is required to cooperate with patch size $l_{patch}$ to guarantee the compatibility between adjacent patches. Considering the computational cost, in this paper we empirically set

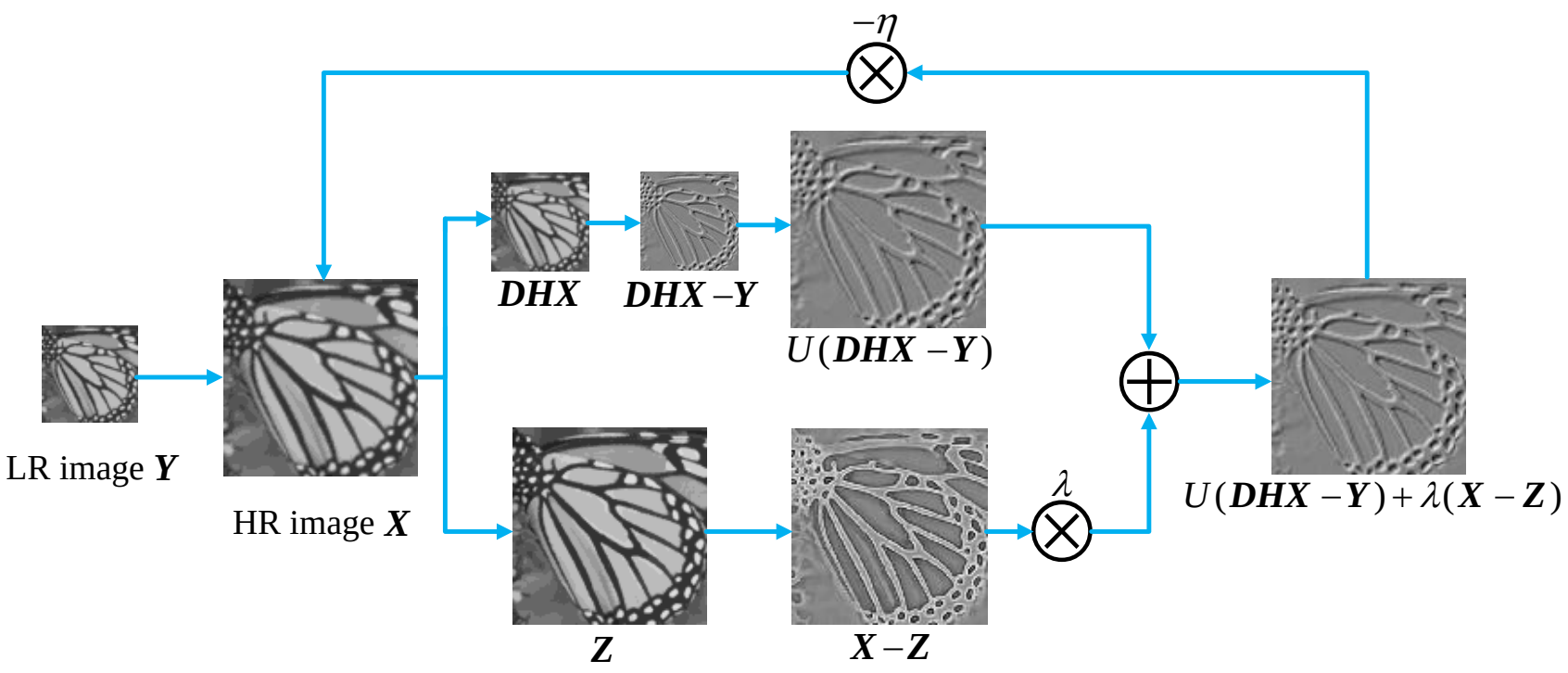

Fig.5 Overall SR framework.

$r_{patch} = 1$ in LR space for efficiency, namely $r_{patch} = k$ in HR space.

### C. Relationship to other edge enhancement methods

As a generic edge-based prior for super-resolution, the proposed sigmoid transformation performs similarly with other edge enhancement methods. In this part, we discuss the relationship and inherent difference between the proposed sigmoid transformation and other representative edge enhancement methods.

Fattal [11] proposed an edge statistics prior for image upscaling, namely edge-frame continuity modulus (EFCM), which is learned to characterize the marginal distribution of the gradients over the whole image. However, the imposed constrain on local continuity measures may not fit varying types of edges and generate jaggy artifacts, besides the computational cost is considerable as reported in [11].

Sun [13][16] establish a generic prior called gradient profile prior, and proposed gradient field transformation to transfer image gradient field guided by the prior knowledge learned from large training sets. However, the gradient profile only utilizes the pixel information along the gradient direction with local image structure neglected, leading to susceptibility to noises and undulation.

Shock filter is another representative category of edge enhancement approach, which is commonly designed to enhance edges detected by edge detectors. Recently, shock filter has been introduced into SR applications [] performing as an edge enhancement operation. However, as shock filter is sensitive to noise, the noise is supposed to be amplified while enhancing the edges.

Compared with other edge enhancement methods, the proposed sigmoid transformation performs twofold distinct differences or advantages. Firstly, the sigmoid transformation performs patch-wise and slope-based, namely the sharpening operation is based on local image structure, which is more reasonable with better suppression of noise. Secondly, our method performs superior efficiency, which is further demonstrated in **Section IV**.

## III. SISR FRAMEWORK BASED ON SIGMOID TRANSFORMATION

In our scenario, the SISR process is realized through typical iterative reconstruction [21][27], where the patch-wise sigmoid transformation performs as an imposed sharpening regularization cooperating with the reconstruction errors.

Given an HR image $\boldsymbol{X}$ and a corresponding LR image $\boldsymbol{Y}$, the degradation process can be formulated as

$$\boldsymbol{Y} = \boldsymbol{DHX} + \boldsymbol{N} \tag{8}$$

where $\boldsymbol{H}$ is the blurring operator, $\boldsymbol{D}$ determines the decimation matrix and $\boldsymbol{N}$ represents the additional Gaussian noise in $\boldsymbol{Y}$. To realize the restoration of HR image $\boldsymbol{X}$, we integrate the patch-wise sigmoid transformation as a sharpening regularization term with the reconstruction error in the cost function as

$$\boldsymbol{X} = argmin_{\boldsymbol{X}}\{\|\boldsymbol{DHX} - \boldsymbol{Y}\|^2 + \lambda \, \|\boldsymbol{X} - \boldsymbol{Z}\|^2\} \tag{9}$$

where $\boldsymbol{Z}$ serves as a sharpened image of $\boldsymbol{X}$ utilizing the proposed patch-wise sigmoid transformation.

To solve the minimization problem in (9), we refer to our previous work in [] and utilize the fast upscaling technique in our SISR framework

$$\boldsymbol{X}^{l+1} = \boldsymbol{X}^{l} - \eta(U(\boldsymbol{DHX}^{l} - \boldsymbol{Y}) + \lambda\,(\boldsymbol{X}^{l} - \boldsymbol{Z}^{l})) \tag{10}$$

where $\boldsymbol{X}^{l}$, $\boldsymbol{X}^{l+1}$ are estimators of HR image $\boldsymbol{X}$ in $l^{th}$ and $l+1^{th}$ iteration respectively, $\lambda$ represents the regularization parameter weighting the regularization cost against the reconstruction error, $\eta$ serves as the stepsize and $U(\cdot)$ refers to the upscaling technique proposed in [37].

The overall framework is shown in **Fig. 5** and further summarized in algorithm 1.

***Algorithm 1: SISR based on sigmoid transformation***

***Input:*** LR image $\boldsymbol{Y}$, scaling factor $\gamma$

***Initialize:*** Upscale the LR image $\boldsymbol{Y}$ utilizing interpolation method with bicubic spline to obtain original HR image $\boldsymbol{X}^0$

***Do:***

1) Re-degenerate the HR image $\boldsymbol{X}^{l-1}$ and compute the reconstruction error $(\boldsymbol{DHX}^{l-1} - \boldsymbol{Y})$, then utilize the upscaling technique [37] to upscale the reconstruction error;

2) Compute the sharpened image $\boldsymbol{Z}^{l-1}$ based on $\boldsymbol{X}^{l-1}$ utilizing the proposed patch-wise sigmoid transformation, then derive the difference between $\boldsymbol{X}^{l-1}$ and $\boldsymbol{Z}^{l-1}$ as a regularization term;

3) Update $\boldsymbol{X}^{l-1}$ to derive $\boldsymbol{X}^{l}$ according to (10).

***Until:*** Stopping criteria are satisfied

***Output:*** Reconstructed HR image $\boldsymbol{X}$

## IV. EXPERIMENTAL RESULTS

***Implementation:*** As human vision is more sensitive to brightness change, we only apply our method on brightness channel (Y) with color channels (UV) upscaled by bicubic interpolation for color images. In our experiments, the HR images are firstly blurred by a $7 \times 7$ Gaussian kernel with $\sigma = 1.2$, and then dwonsampled with factor $k = 3$ to serve as the input LR images. During the implementation of our method, regularization parameter $\lambda$ and learning rate $\eta$ are set to be 0.2 and 0.1 respectively, while the maximum number of iterations specified as 30.

All the experiments are coded in Matlab R2011 and running on a workstation with Septuple Core i7 @ 3.60 GHz CPUs and 16GB RAM.

***Metrics:*** To evaluate and compare the results quantitatively, peak-signal-to-noise ratio (PSNR) and mean structure similarity (SSIM) are utilized as metrics, which are defined as

$$\begin{cases} PSNR = 10log_{10}\left(\frac{255^2}{MSE}\right), MSE = \frac{1}{MN}\sum_{i=1}^{M}\sum_{j=1}^{N}(\boldsymbol{Y}(i,j) - \boldsymbol{X}(i,j))^2 \\ SSIM = \frac{(2\mu_x\mu_y+c_1)(2\sigma_x\sigma_y+c_2)}{(\mu_x{}^2+\mu_y{}^2+c_1)(\sigma_x{}^2+\sigma_y{}^2+c_2)}, \begin{cases} c_1 = (k_1L)^2 \\ c_2 = (k_2L)^2 \end{cases} \end{cases}, \quad (11)$$

where $\mu_x, \mu_y$ are mean value of image $\boldsymbol{X}$ and $\boldsymbol{Y}$ respectively, $\sigma_x, \sigma_y$ are standard variance of $\boldsymbol{X}$ and $\boldsymbol{Y}$ respectively, $c_1, c_2$ are two stabilizing constants with $L$ representing the dynamics of a pixel value, and $k_1, k_2$ are generally set to be 0.01 and 0.03 respectively.

### *A. Experimental analysis of parameters*

In this section, we focus on the sharpness parameter $K$ and location parameter $B$, conduct extensive experiments on **Set5** and **Set14** datasets to test the effects of our method with various parameter settings.

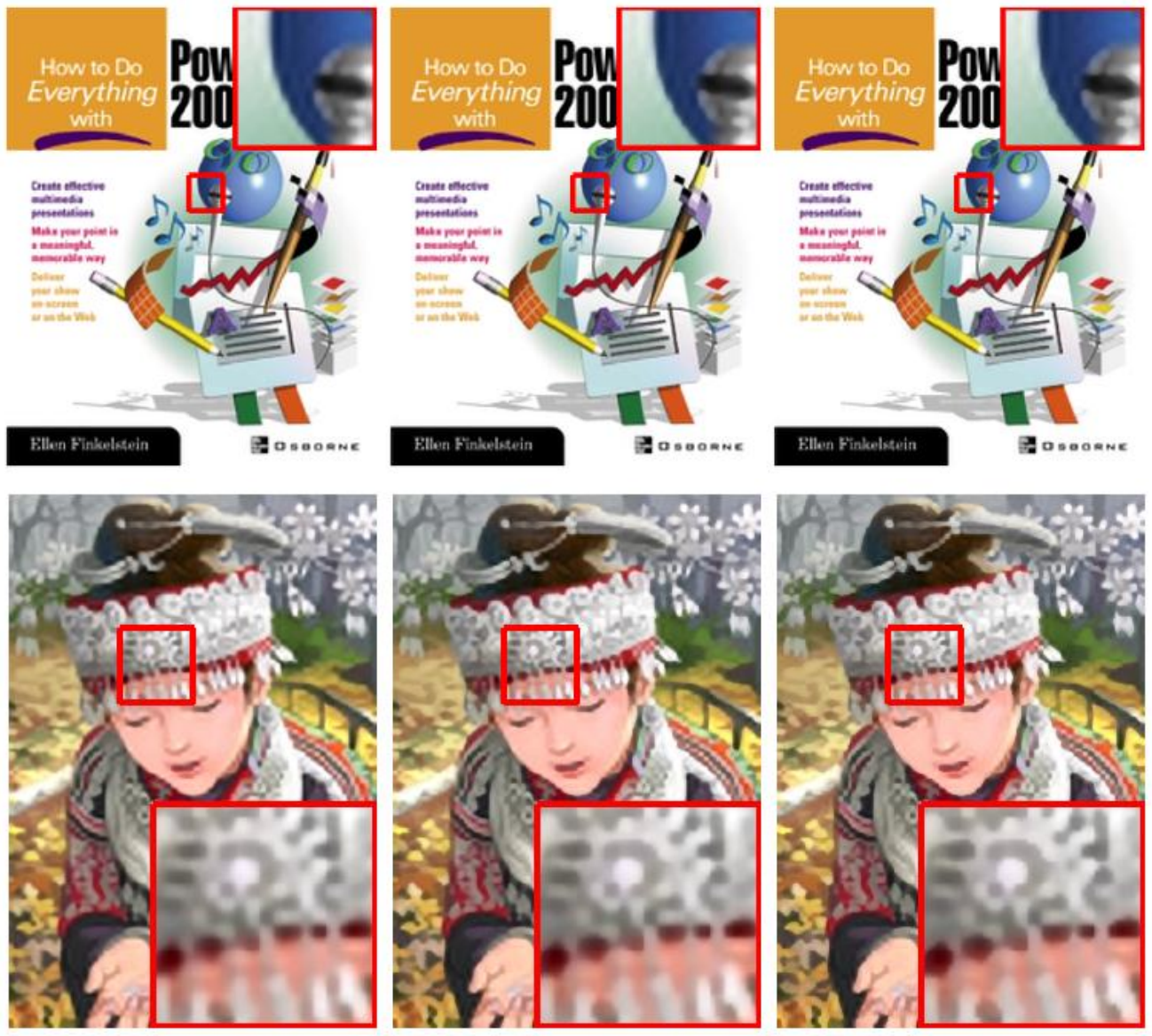

(a) K=2 (b) K=3 (c) K=4

Fig.6 Effect of parameter $K$ on image *PPT* and *Comic*. The images are reconstruction results (3X) with $K = 2,3,4$.

**Fig. 6** presents the effects of sharpness parameter $K$ on the reconstruction result. As we can see, larger $K$ leads to sharper and narrower edges, however, over-sharp artifacts can be visible when $K$ is over-large.

In **Fig. 7**, effects of location parameter $B$ on the reconstruction result is given. Zooming in on the images, the displacement of edges can be observed between the results with different values of $B$. Compare the enlarged region of *Lenna*, the widening of the brown region under Lenna's hat with $B$ increased can be noticeable.

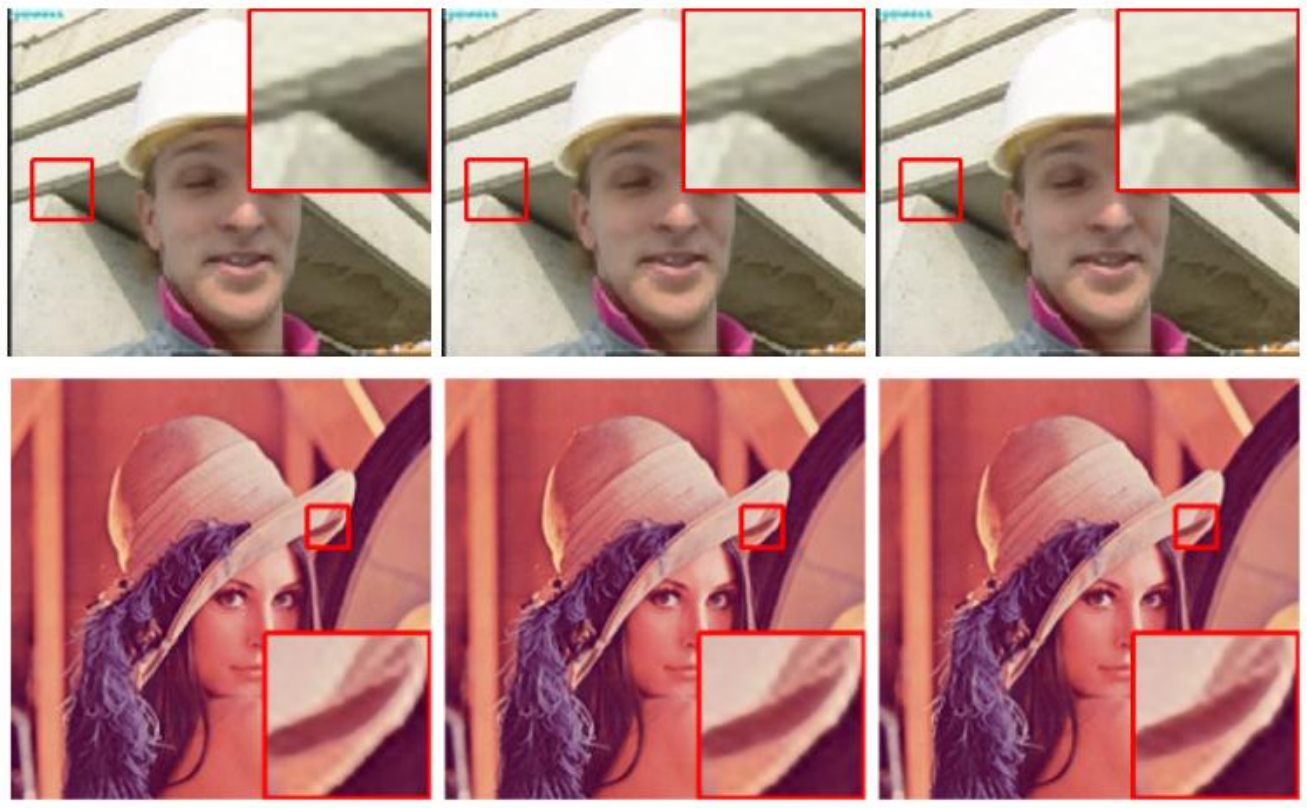

(a) B=-1 (b) B=0 (c) B=1

Fig.7 Effect of parameter $B$ on image *Foreman* and *Lenna*. The images are reconstruction results (3X) with $B = -1,0,1$.

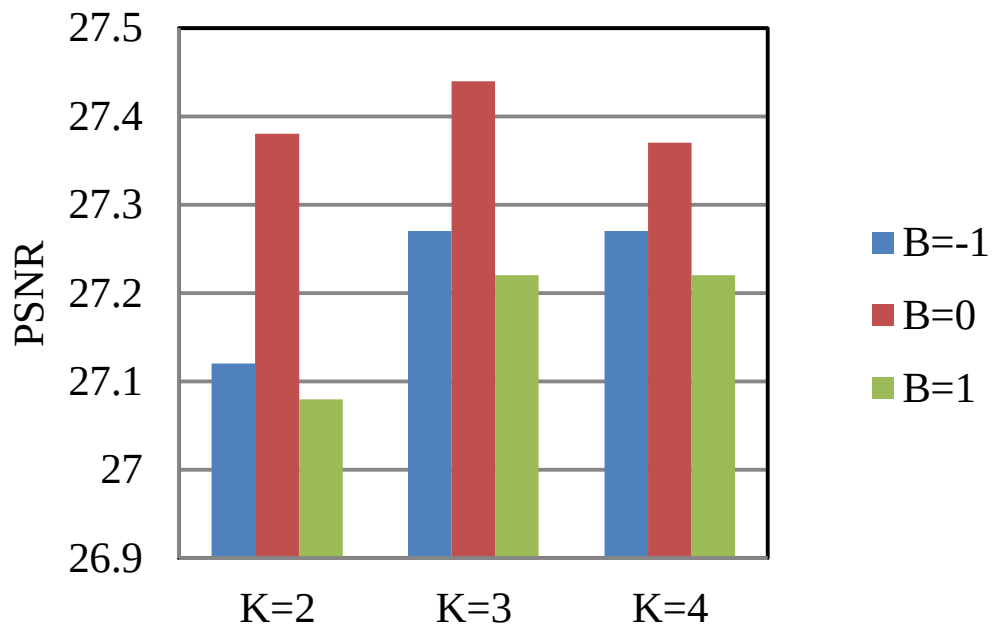

Fig.8 Average PSNR of different parameter settings.

Comparison of quantitative performance is exhibited in **Table. I** with average performance presented in **Fig. 8**. As sharpness parameter $K$ increases, the reconstruction performance is first improved and then begins to deteriorate due to the over-sharp effects. Concerning location parameter $B$, obviously it has great impacts on the performance with $B = 0$ performing as the dominant one. However, note that it does not mean no benefits of $B$ in fine-tuning the locations of edges, which can be validated by some superior results in **Table. I** with $B \neq 0$. As edges within an image varies remar kably, the fine-tuning is required to implement adaptively, nevertheless, the global selection of $B$ and implementation of fine-tuning in our scenario may not adapt to most edges, commonly leading to degeneration of reconstruction performance.

Note that the adaptive selection of $K$ and $B$ is not the focus of this paper, we empirically set $K=2$ and $B=0$ for following experiments and leave this problem for further investigation.

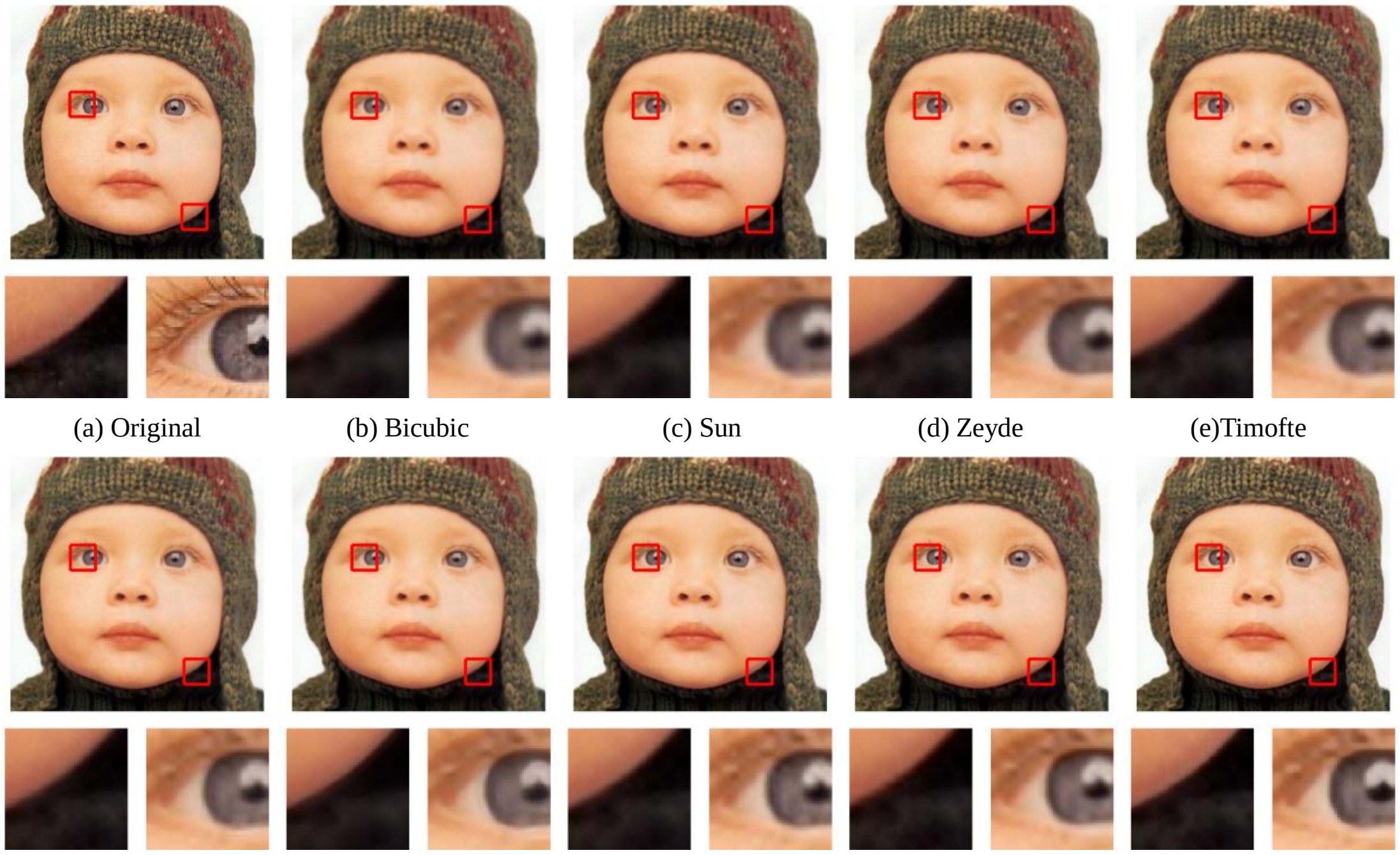

(a) Original (b) Bicubic (c) Sun (d) Zeyde (e)Timofte

(f) Yang (g) Kim (h) Yang (i) Dong (j)Proposed

Fig.9 SISR results (3X) on image *Baby* by ranged methods.

TABLE I COMPARISON OF PSNR WITH DIFFERENT PARAMETER SETTINGS. THE BEST RESULT ARE SHOWN IN BOLD.

| | $K=2$ | | | $K=3$ | | | $K=4$ | | |
|---|---|---|---|---|---|---|---|---|---|
| | $B=-1$ | $B=0$ | $B=1$ | $B=-1$ | $B=0$ | $B=1$ | $B=-1$ | $B=0$ | $B=1$ |
| **Baby** | 33.00 | **33.22** | 32.93 | 32.92 | 33.04 | 32.87 | 32.72 | 32.79 | 32.69 |
| **Bird** | 30.83 | **31.19** | 30.96 | 30.85 | 31.08 | 30.94 | 30.73 | 30.88 | 30.78 |
| **Butterfly** | 23.97 | 24.75 | 24.02 | 24.68 | 25.21 | 24.67 | 24.90 | **25.25** | 24.88 |
| **Head** | 30.37 | **30.41** | 30.32 | 30.36 | 30.36 | 30.29 | 30.30 | 30.30 | 30.24 |
| **Woman** | 28.70 | 28.79 | 27.99 | **28.94** | 28.90 | 28.26 | 28.88 | 28.78 | 28.31 |
| **Baboon** | 21.28 | **21.32** | 21.31 | 21.27 | 21.30 | 21.29 | 21.25 | 21.27 | 21.27 |
| **Barbara** | 25.09 | **25.14** | 25.05 | 25.09 | 25.12 | 25.06 | 25.06 | 25.07 | 25.03 |
| **Coastguard** | 25.56 | 25.73 | 25.70 | 25.60 | **25.75** | **25.75** | 25.60 | 25.73 | 25.75 |
| **Comic** | 22.22 | 22.33 | 22.08 | 22.35 | **22.38** | 22.18 | 22.35 | 22.34 | 22.19 |
| **Flowers** | 25.70 | 25.98 | 25.80 | 25.84 | **26.04** | 25.92 | 25.86 | 26.00 | 25.92 |
| **Foreman** | 31.05 | 31.11 | 30.20 | **31.35** | 31.28 | 30.56 | 31.37 | 31.25 | 30.66 |
| **Lena** | 30.34 | **30.63** | 30.49 | 30.39 | 30.58 | 30.50 | 30.33 | 30.46 | 30.42 |
| **Monarch** | 29.17 | 29.75 | 29.12 | 29.68 | **30.05** | 29.57 | 29.80 | 30.03 | 29.68 |
| **Pepper** | 30.29 | **30.48** | 30.18 | 30.38 | **30.48** | 30.26 | 30.34 | 30.38 | 30.23 |
| **PPT** | 23.00 | 23.45 | 23.27 | 23.30 | **23.63** | 23.53 | 23.41 | 23.64 | 23.59 |
| **Zebra** | 26.37 | 27.07 | 26.79 | 26.58 | **27.11** | 27.02 | 26.56 | 26.95 | 26.95 |
| **Bridge** | 24.84 | **24.99** | 24.90 | 24.88 | **24.99** | 24.94 | 24.87 | 24.94 | 24.92 |
| **Man** | 26.30 | 26.49 | 26.28 | 26.42 | **26.55** | 26.40 | 26.44 | 26.52 | 26.41 |
| **Average** | 27.12 | 27.38 | 27.08 | 27.27 | **27.44** | 27.22 | 27.27 | 27.37 | 27.22 |

*B. Comparison to other methods*

In this section, experiments are conducted on Set5 and Set14 datasets to compare the reconstruction performance of other state-of-the-art approaches, including bicubic, Kim *et al.*'s method [14], Zeyde *et al.*'s method [31], Timofte *et al.*'s method [18] and Yang *et al.*'s method [28], Dong *et al.*'s method [36], Sun *et al.*'s method [16] and Yang *et al.*'s method [17]. The source codes of approaches [14][31][18][28][17][36] are downloaded from the authors'homepages, and we refer to Yang's implementation work in [19] to derive other approaches [10][13][16] for no available originally released codes. Recommended parameters of comparing methods by the authors are used in our experiments.

We summarize the parameter settings for our method in **Table. II**, and the reconstruction results are exhibited in **Figs. 9-13** with quantitative results presented in **Table. III**.

TABLE II PARAMETER SETTINGS FOR THE PROPOSED METHOD.

| **Parameters** | **Values** |
|---|---|
| Sharpness $K$ | 2 |
| Location $B$ | 0 |
| Patch size $l_{patch}$ | 3 (in LR space) |
| stride $r_{patch}$ | 1 (in LR space) |

- Visual quality

From the reconstruction results on image *Baby*, *Butterfly*, *Head*, *Comic* and *Lena* with upscaling factor 3 shown in **Figs. 9-13**, we can see that the bicubic interpolation method blurs the edge and texture regions remarkably with a serious loss of image de tails. Sun's method generates sharper edges, however the blurring effect is still noticeable. For Zeyde's method, Timofte's method, Yang's method and Yang's method, the blurring effect is further alleviated, but some ringing artifacts around the edges can still be visible. Relying on adaptive sparse representation, Dong's method achieves relatively excellent visual quality, however some fine details like texture are missed due to the sparsity regularization. Kim's method generates sharp edges and distinct details, performs superior reconstruction performance. Concerning our proposed method, comparable top visual quality is achieved with even more fine details reconstructed, demonstrating the superiority of our proposed method.

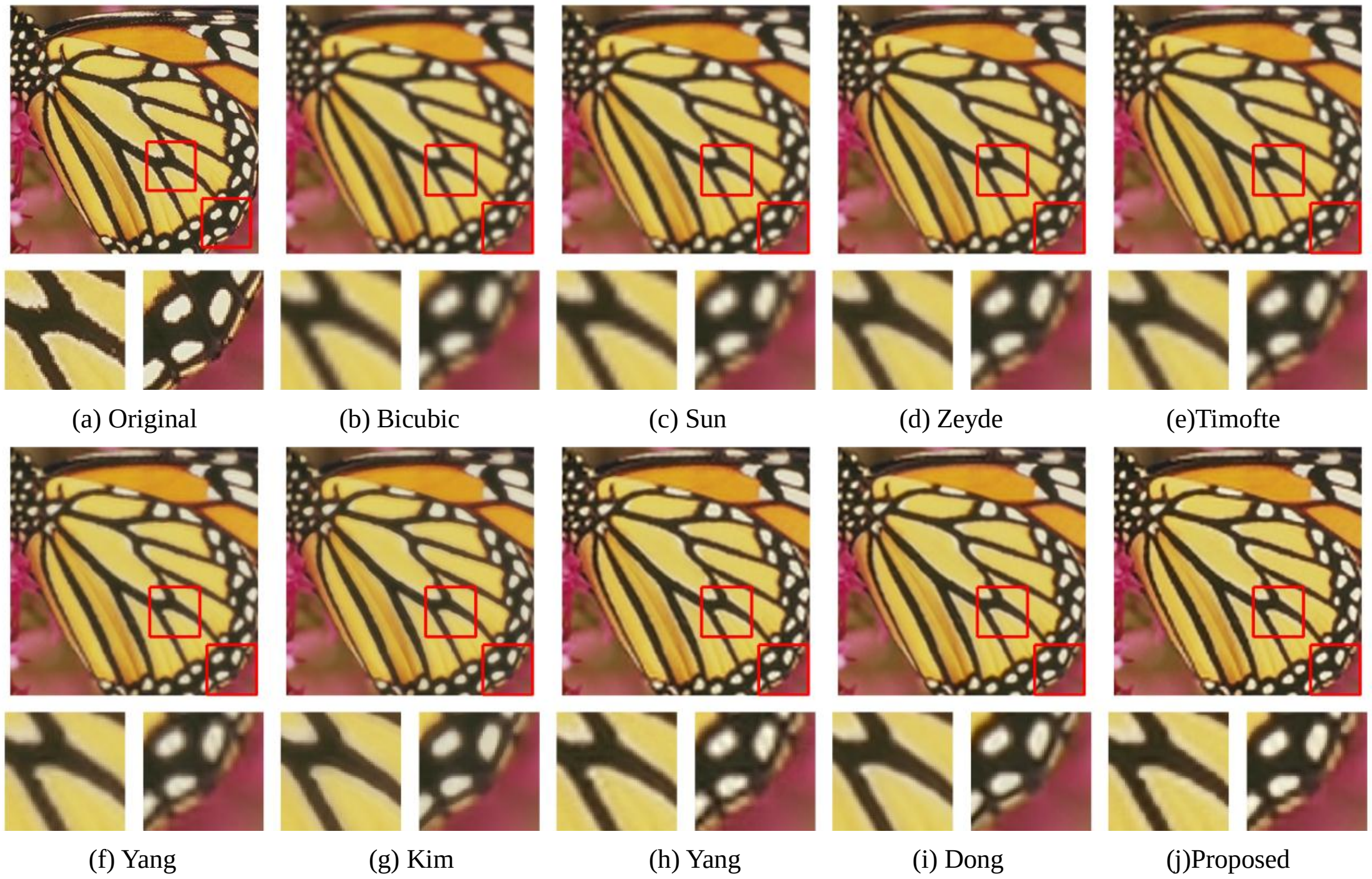

Fig.10 SISR results (3X) on image *Butterfly* by ranged methods.

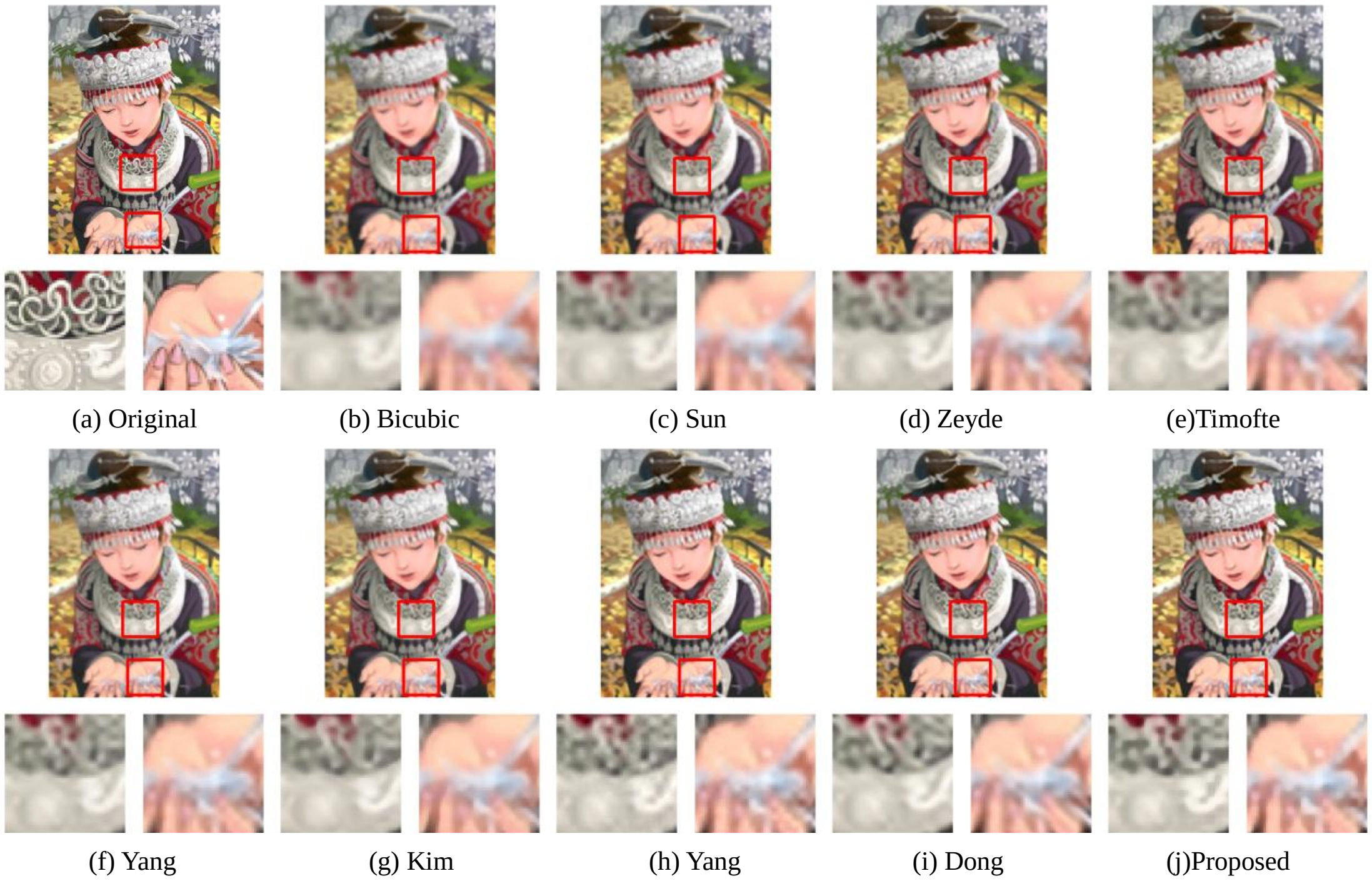

Fig.11 SISR results (3X) on image *Comic* by ranged methods.

● Quantitative metrics

As we can see in **Table.** III, the proposed method outperforms other approaches on all the test images with highest PSNR and SSIM values. Compared with Sun's method, our method performs remarkable improvement with PSNR value improved by 1.61 dB in average. Concerning state-of-the-art Kim's and Yang's method, the proposed method improves the average PSNR value with about 0.7 and 0.3 dB respectively, demonstrating the superior reconstruction performance of the proposed method.

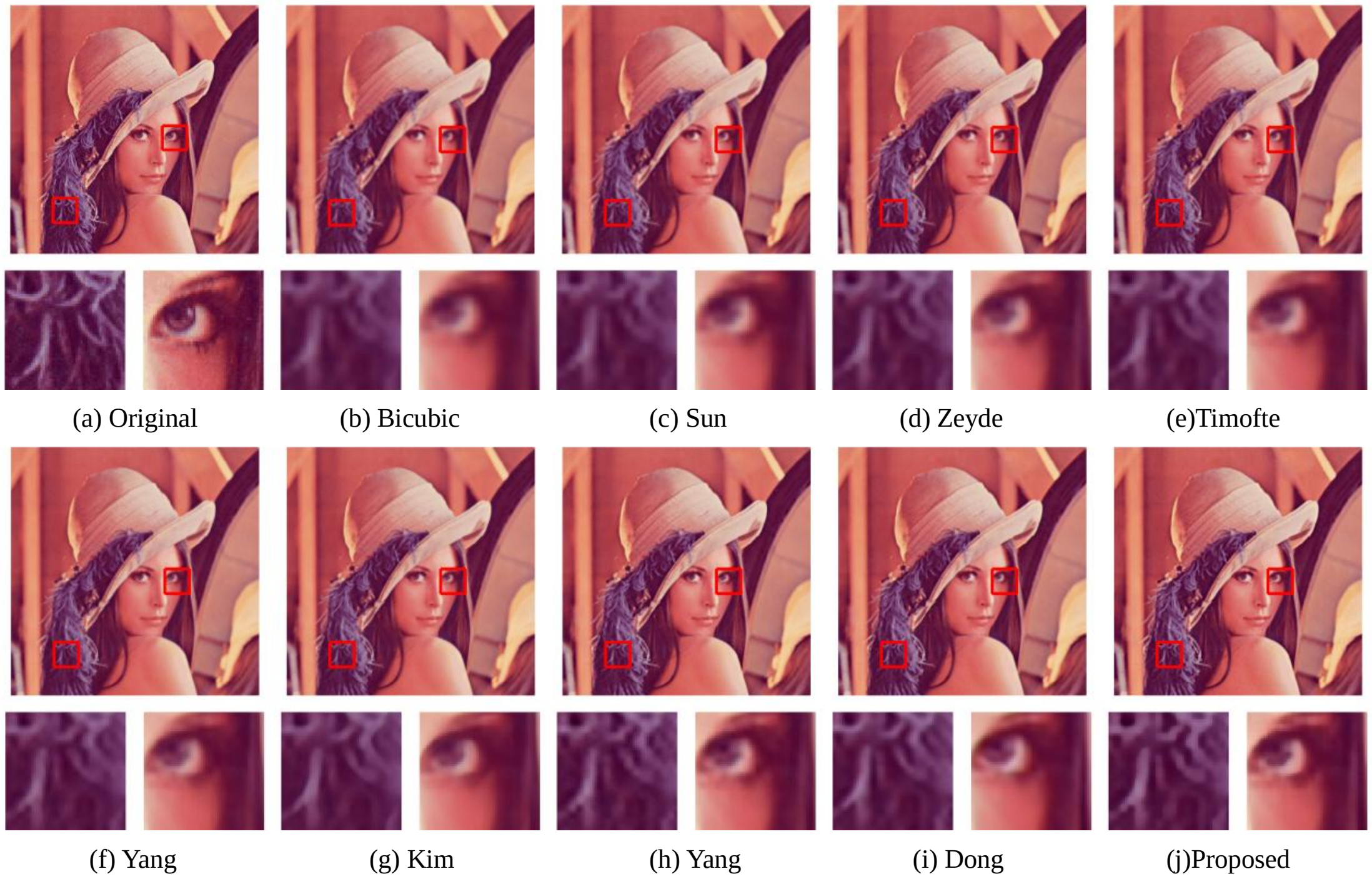

Fig.12 SISR results (3X) on image *Lena* by ranged methods.

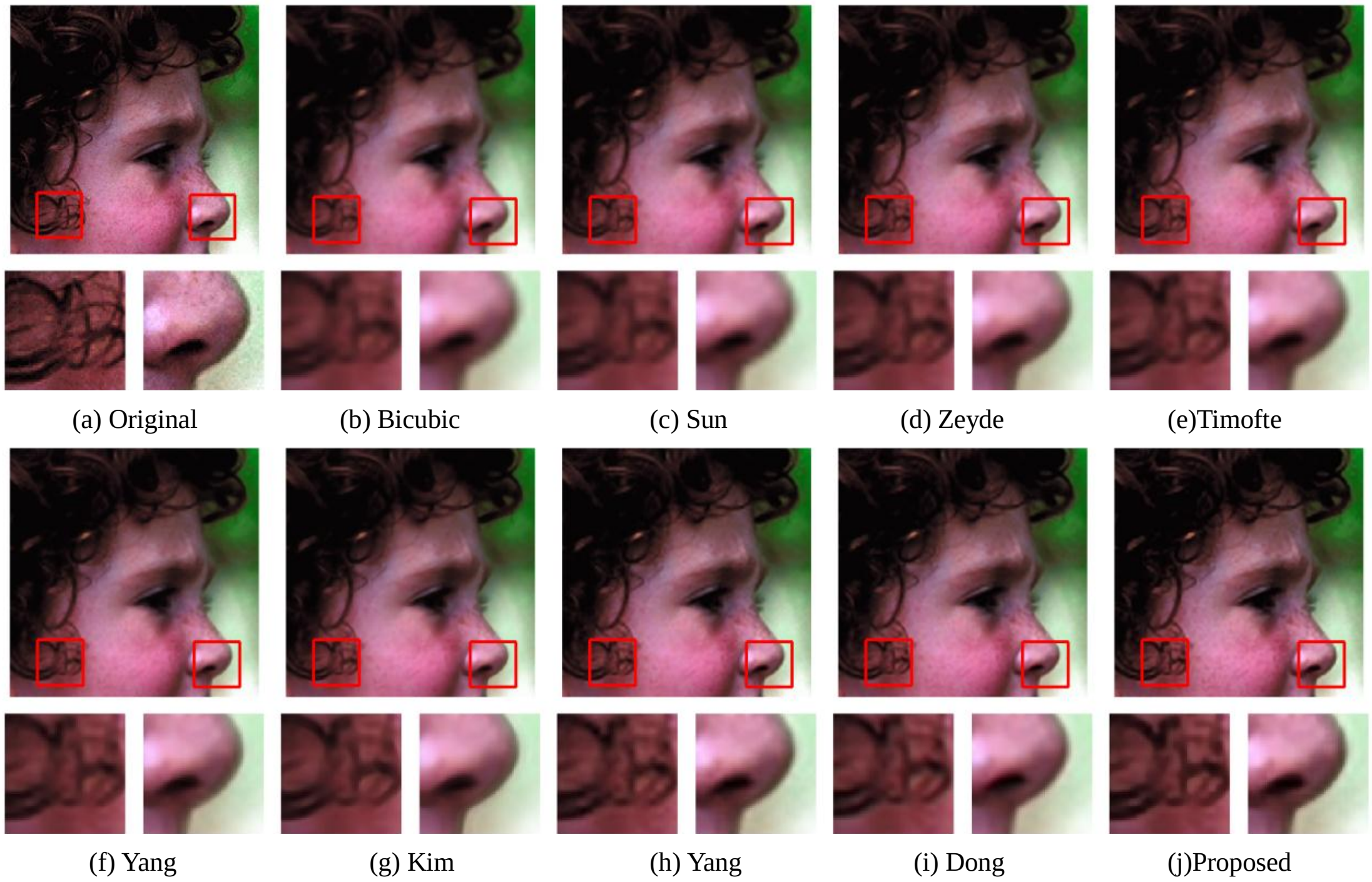

Fig.13 SISR results (3X) on image *Head* by ranged methods.

● Running time

In terms of processing efficiency, obviously the running time of bicubic interpolation method is much faster for no complex calculations. Some dictionary-based methods like Zeyde's, Timofte's and Yang's methods also performs superior efficiency, however note that the learning processes of these methods are hugely time-consuming. Concerning our proposed method, there is no need for additional training process and the running time is still faster than Kim's, Yang's, Yang's, Dong's and Sun's methods, even comparable to

TABLE III COMPARISON OF PSNR, SSIM AND RUNNING TIME. THE BEST RESULT ARE SHOWN IN BOLD.

| | Metrics | Bicubic | Sun [16] | Zeyde [31] | Timofte [18] | Yang [28] | Kim [14] | Yang [17] | Dong [36] | Proposed |
|---|---|---|---|---|---|---|---|---|---|---|
| **Baby** | PSNR | 30.91 | 31.47 | 32.12 | 32.11 | 32.11 | 32.22 | 32.86 | 28.32 | **33.22** |
| | SSIM | 0.848 | 0.861 | 0.871 | 0.873 | 0.874 | 0.875 | 0.886 | 0.825 | **0.895** |
| | Time | **0.011** | 59.354 | 2.779 | 0.804 | 192.766 | 18.172 | 7.740 | 400.262 | 5.165 |
| **Bird** | PSNR | 28.72 | 29.29 | 30.06 | 30.06 | 30.00 | 30.29 | 30.89 | 26.26 | **31.19** |
| | SSIM | 0.870 | 0.882 | 0.894 | 0.896 | 0.895 | 0.899 | 0.903 | 0.815 | **0.904** |
| | Time | **0.005** | 14.287 | 1.172 | 0.526 | 56.172 | 9.271 | 3.558 | 122.515 | 1.902 |
| **Butterfly** | PSNR | 21.33 | 21.95 | 22.93 | 22.86 | 22.96 | 23.76 | 23.88 | 20.23 | **24.75** |
| | SSIM | 0.742 | 0.769 | 0.815 | 0.810 | 0.817 | 0.846 | 0.835 | 0.759 | **0.849** |
| | Time | **0.005** | 10.342 | 0.935 | 0.543 | 42.526 | 18.601 | 2.844 | 134.362 | 1.663 |
| **Head** | PSNR | 29.43 | 29.67 | 29.95 | 30.00 | 30.00 | 30.08 | 30.25 | 28.08 | **30.41** |
| | SSIM | 0.692 | 0.703 | 0.714 | 0.717 | 0.718 | 0.718 | 0.727 | 0.670 | **0.741** |
| | Time | **0.005** | 14.094 | 1.033 | 0.473 | 55.657 | 50.75 | 2.889 | 109.330 | 1.829 |
| **Woman** | PSNR | 25.66 | 26.24 | 27.24 | 27.20 | 27.22 | 27.56 | 28.33 | 23.46 | **28.79** |
| | SSIM | 0.840 | 0.855 | 0.874 | 0.875 | 0.875 | 0.879 | 0.892 | 0.817 | **0.899** |
| | Time | **0.005** | 13.434 | 1.112 | 0.449 | 53.079 | 9.756 | 3.080 | 119.914 | 1.776 |
| **Baboon** | PSNR | 20.75 | 20.89 | 21.03 | 21.08 | 21.09 | 21.13 | 21.24 | 20.37 | **21.32** |
| | SSIM | 0.423 | 0.441 | 0.457 | 0.464 | 0.468 | 0.470 | 0.490 | 0.423 | **0.512** |
| | Time | **0.010** | 72.780 | 2.641 | 0.834 | 165.652 | 26.970 | 9.861 | 398.111 | 4.460 |
| **Barbara** | PSNR | 24.04 | 24.27 | 24.89 | 24.60 | 24.63 | 24.68 | 24.90 | 22.86 | **25.14** |
| | SSIM | 0.678 | 0.692 | 0.710 | 0.712 | 0.714 | 0.715 | 0.729 | 0.652 | **0.743** |
| | Time | **0.014** | 116.127 | 4.622 | 1.161 | 297.839 | 33.783 | 15.519 | 699.157 | 7.505 |
| **Coastguard** | PSNR | 24.60 | 24.85 | 25.26 | 25.21 | 25.35 | 25.34 | 25.54 | 23.93 | **25.73** |
| | SSIM | 0.527 | 0.548 | 0.569 | 0.574 | 0.582 | 0.577 | 0.596 | 0.518 | **0.627** |
| | Time | **0.006** | 23.299 | 1.241 | 0.547 | 70.530 | 12.450 | 3.957 | 156.050 | 2.121 |
| **Comic** | PSNR | 20.73 | 21.05 | 21.49 | 21.57 | 21.62 | 21.76 | 22.04 | 19.72 | **22.33** |
| | SSIM | 0.615 | 0.640 | 0.672 | 0.677 | 0.682 | 0.689 | 0.707 | 0.596 | **0.731** |
| | Time | **0.005** | 16.556 | 1.210 | 0.471 | 64.117 | 14.991 | 4.115 | 162.344 | 1.978 |
| **Flowers** | PSNR | 24.25 | 24.62 | 25.15 | 25.18 | 25.21 | 25.35 | 25.68 | 22.96 | **25.98** |
| | SSIM | 0.728 | 0.743 | 0.760 | 0.762 | 0.764 | 0.766 | 0.774 | 0.696 | **0.782** |
| | Time | **0.008** | 36.631 | 2.259 | 0.799 | 131.354 | 22.832 | 6.917 | 289.215 | 3.458 |
| **Foreman** | PSNR | 28.23 | 28.81 | 29.75 | 29.71 | 29.63 | 30.30 | 30.82 | 27.11 | **31.11** |
| | SSIM | 0.863 | 0.874 | 0.889 | 0.830 | 0.891 | 0.898 | 0.902 | 0.860 | **0.906** |
| | Time | **0.008** | 18.407 | 1.242 | 0.495 | 67.332 | 9.451 | 3.659 | 158.952 | 2.101 |
| **Lena** | PSNR | 28.81 | 29.23 | 29.84 | 29.86 | 29.88 | 30.14 | 30.49 | 26.45 | **30.63** |
| | SSIM | 0.755 | 0.766 | 0.777 | 0.779 | 0.780 | 0.783 | 0.790 | 0.726 | **0.797** |
| | Time | **0.010** | 55.387 | 2.790 | 0.827 | 192.144 | 17.336 | 8.212 | 410.589 | 5.184 |
| **Monarch** | PSNR | 26.63 | 27.22 | 28.07 | 28.04 | 28.10 | 28.70 | 28.95 | 25.17 | **29.75** |
| | SSIM | 0.880 | 0.890 | 0.904 | 0.903 | 0.905 | 0.914 | 0.912 | 0.866 | **0.917** |
| | Time | **0.015** | 91.493 | 4.047 | 1.167 | 287.476 | 37.932 | 10.566 | 652.745 | 7.074 |
| **Pepper** | PSNR | 29.06 | 29.43 | 30.03 | 29.93 | 29.95 | 30.16 | 30.39 | 26.65 | **30.48** |
| | SSIM | 0.764 | 0.772 | 0.779 | 0.779 | 0.779 | 0.781 | 0.786 | 0.732 | **0.788** |
| | Time | **0.011** | 58.071 | 2.850 | 0.881 | 190.785 | 17.517 | 8.407 | 404.304 | 5.232 |
| **PPT** | PSNR | 21.17 | 21.56 | 22.44 | 22.30 | 22.30 | 22.65 | 23.13 | 19.76 | **23.45** |
| | SSIM | 0.830 | 0.846 | 0.870 | 0.865 | 0.865 | 0.873 | **0.886** | 0.812 | **0.886** |
| | Time | **0.010** | 47.589 | 3.078 | 0.990 | 176.978 | 42.968 | 9.834 | 667.489 | 6.449 |
| **Zebra** | PSNR | 23.55 | 24.20 | 25.22 | 25.20 | 25.17 | 25.71 | 26.45 | 21.31 | **27.07** |
| | SSIM | 0.704 | 0.732 | 0.762 | 0.766 | 0.770 | 0.771 | 0.804 | 0.667 | **0.826** |
| | Time | **0.009** | 54.000 | 2.460 | 0.876 | 167.638 | 43.599 | 10.067 | 398.469 | 4.222 |
| **Bridge** | PSNR | 23.65 | 23.94 | 24.33 | 24.35 | 24.38 | 24.42 | 24.77 | 22.63 | **24.99** |
| | SSIM | 0.585 | 0.608 | 0.633 | 0.639 | 0.645 | 0.642 | 0.672 | 0.566 | **0.698** |
| | Time | **0.010** | 59.635 | 2.859 | 0.957 | 195.138 | 34.385 | 11.670 | 459.461 | 5.221 |
| **Man** | PSNR | 24.82 | 25.16 | 25.72 | 25.73 | 25.78 | 25.96 | 26.26 | 23.46 | **26.49** |
| | SSIM | 0.676 | 0.695 | 0.718 | 0.721 | 0.715 | 0.725 | 0.745 | 0.654 | **0.763** |
| | Time | **0.010** | 67.478 | 2.791 | 0.937 | 178.226 | 31.548 | 10.812 | 427.198 | 5.236 |
| **Average** | PSNR | 25.32 | 25.77 | 26.42 | 26.39 | 26.41 | 26.68 | 27.05 | 22.82 | **27.38** |
| | SSIM | 0.723 | 0.740 | 0.759 | 0.758 | 0.763 | 0.768 | 0.780 | 0.703 | **0.792** |
| | Time | **0.009** | 46.054 | 2.285 | 0.763 | 143.634 | 25.128 | 7.428 | 342.804 | 4.032 |

Zeyde's method, demonstrating the superior efficiency for the proposed simple sigmoid transformation.

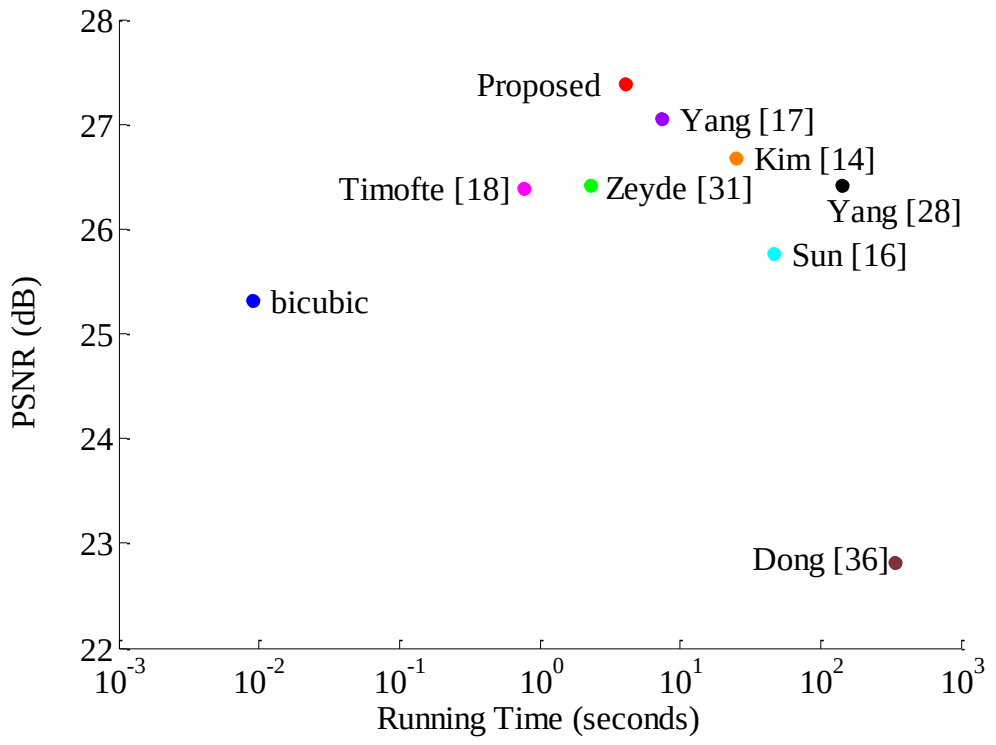

Fig.14 Plot of the trade-off between accuracy and speed for ranged methods with upscaling factor 3. The results present the mean PSNR values and running time on the test images.

Overall, as shown in **Fig. 14**, our method realizes dominant reconstruction performance without much loss of efficiency, performing superior to Yang's method known as state-of-the-art SISR approach with respect to PSNR value, and competitive to Zeyde's methods concerning running time. As our method requires no training process, the superiority in efficiency can be more prominent if we take training time into consideration.

### *C. Robustness*

To further demonstrate the robustness of the proposed method to various Gaussian blurring kernels, upscaling factors and noise intensities, additional experiments are conducted on Set5 and Set14 datasets in this section, and the quantitative results are presented in **Tables. IV-VI**.

- Blurring kernel

**Table. IV** presents the comparison of reconstruction performance with various blurring kernels under the condition of upscaling factor 3. As we can see, the overall reconstruction performance for various methods degenerates with blurring kernel width increased. Under the condition of $\sigma = 0.8$, although Kim's method performs superiorly with highest PSNR value in average, our method is still comparable and outperforms other approaches. Concerning conditions of $\sigma = 1.2$ and $\sigma = 1.6$, the proposed method outperforms other approaches in average while Yang's method, as the most competitive one, also performs competitively when $\sigma = 1.6$. Overall, with blurring kernel width varying, the proposed method performs as the dominant one on most test images when $\sigma = 1.2$ and $\sigma = 1.6$, and achieves second best performance after Kim's method when $\sigma = 0.8$, indicating

TABLE IV COMPARISON OF RECONSTRUCTION PERFORMANCE WITH VIROUS $\sigma$. THE BEST RESULTS ARE SHOWN IN BOLD.

| | $\sigma$ | Bicubic | Sun [16] | Zeyde [31] | Timofte [18] | Yang [28] | Kim [14] | Yang [17] | Dong [36] | Proposed |
|---|---|---|---|---|---|---|---|---|---|---|
| | 0.8 | 31.73 | 32.07 | 33.02 | 33.04 | 33.00 | 33.12 | 32.67 | 28.31 | **33.21** |
| **Baby** | 1.2 | 30.91 | 31.47 | 32.12 | 32.11 | 32.11 | 32.22 | 32.86 | 28.32 | **33.22** |
| | 1.6 | 30.16 | 30.78 | 31.17 | 31.12 | 31.08 | 31.18 | 32.89 | 28.28 | **32.90** |
| | 0.8 | 29.75 | 30.12 | 31.35 | 31.39 | 31.25 | **31.67** | 30.95 | 26.38 | 31.46 |
| **Bird** | 1.2 | 28.72 | 29.29 | 30.06 | 30.06 | 30.00 | 30.29 | 30.89 | 26.26 | **31.19** |
| | 1.6 | 27.80 | 28.40 | 28.82 | 28.78 | 28.71 | 28.90 | **30.68** | 26.12 | 30.55 |
| | 0.8 | 22.11 | 22.53 | 24.01 | 24.00 | 24.15 | **25.17** | 23.72 | 30.25 | 24.60 |
| **Butterfly** | 1.2 | 21.33 | 21.95 | 22.93 | 22.86 | 22.96 | 23.76 | 23.88 | 20.23 | **24.75** |
| | 1.6 | 20.63 | 21.29 | 21.85 | 21.74 | 21.76 | 22.36 | 23.99 | 20.18 | **24.32** |
| | 0.8 | 29.82 | 29.97 | 30.37 | 30.43 | 30.40 | **30.53** | 30.19 | 28.06 | 30.42 |
| **Head** | 1.2 | 29.43 | 29.67 | 29.95 | 30.00 | 30.00 | 30.08 | 30.25 | 28.08 | **30.41** |
| | 1.6 | 29.06 | 29.32 | 29.49 | 29.51 | 29.51 | 29.55 | 30.19 | 28.04 | **30.24** |
| | 0.8 | 26.49 | 26.87 | 28.37 | 28.39 | 28.43 | **28.75** | 28.03 | 23.46 | 28.66 |
| **Woman** | 1.2 | 25.66 | 26.24 | 27.24 | 27.20 | 27.22 | 27.56 | 28.33 | 23.46 | **28.79** |
| | 1.6 | 24.93 | 25.53 | 26.13 | 26.06 | 26.02 | 26.30 | **28.55** | 23.41 | 28.36 |
| | 0.8 | 21.01 | 21.11 | 21.29 | 21.34 | 21.33 | **21.37** | 21.23 | 20.35 | 21.35 |
| **Baboon** | 1.2 | 20.75 | 20.89 | 21.03 | 21.08 | 21.09 | 21.13 | 21.24 | 20.37 | **21.32** |
| | 1.6 | 20.51 | 20.63 | 20.73 | 20.77 | 20.78 | 20.80 | **21.21** | 20.32 | 21.18 |
| | 0.8 | 24.44 | 24.59 | 25.00 | 24.98 | 25.02 | 24.94 | 24.83 | 22.85 | **25.12** |
| **Barbara** | 1.2 | 24.04 | 24.27 | 24.89 | 24.60 | 24.63 | 24.68 | 24.90 | 22.86 | **25.14** |
| | 1.6 | 23.67 | 23.90 | 24.13 | 24.14 | 24.14 | 24.24 | **25.00** | 22.83 | 24.95 |
| | 0.8 | 25.00 | 25.18 | 25.62 | 25.55 | 25.63 | 25.65 | 25.54 | 23.93 | **25.68** |
| **Coastguard** | 1.2 | 24.60 | 24.85 | 25.26 | 25.21 | 25.35 | 25.34 | 25.54 | 23.93 | **25.73** |
| | 1.6 | 24.23 | 24.46 | 24.78 | 24.73 | 24.87 | 24.84 | 25.52 | 23.88 | **25.55** |
| | 0.8 | 21.27 | 21.49 | 22.11 | 22.20 | 22.24 | **22.39** | 21.95 | 19.71 | 22.34 |
| **Comic** | 1.2 | 20.73 | 21.05 | 21.49 | 21.57 | 21.62 | 21.76 | 22.04 | 19.72 | **22.33** |
| | 1.6 | 20.24 | 20.55 | 20.85 | 20.88 | 20.90 | 21.01 | **22.05** | 19.67 | 22.02 |
| | 0.8 | 24.91 | 25.16 | 25.91 | 25.96 | 25.98 | **26.18** | 25.65 | 23.02 | 26.08 |
| **Flowers** | 1.2 | 24.25 | 24.62 | 25.15 | 25.18 | 25.21 | 25.35 | 25.68 | 22.96 | **25.98** |
| | 1.6 | 23.66 | 24.03 | 24.37 | 24.37 | 24.37 | 24.48 | **25.62** | 22.85 | 25.61 |
| | 0.8 | 29.00 | 29.41 | 30.88 | 30.98 | 30.84 | **31.69** | 30.59 | 27.12 | 31.06 |
| **Foreman** | 1.2 | 28.23 | 28.81 | 29.75 | 29.71 | 29.63 | 30.30 | 30.82 | 27.11 | **31.11** |
| | 1.6 | 27.55 | 28.13 | 28.61 | 28.59 | 28.49 | 28.93 | 30.61 | 26.99 | **30.65** |
| | 0.8 | 29.44 | 29.71 | 30.57 | 30.63 | 30.61 | **30.88** | 30.40 | 26.43 | 30.60 |
| **Lena** | 1.2 | 28.81 | 29.23 | 29.84 | 29.86 | 29.88 | 30.14 | 30.49 | 26.45 | **30.63** |
| | 1.6 | 28.24 | 28.68 | 29.07 | 29.04 | 29.04 | 29.23 | **30.48** | 26.43 | 30.37 |
| | 0.8 | 27.43 | 27.82 | 29.12 | 29.14 | 29.23 | **30.02** | 28.79 | 25.20 | 29.67 |
| **Monarch** | 1.2 | 26.63 | 27.22 | 28.07 | 28.04 | 28.10 | 28.70 | 28.95 | 25.14 | **29.75** |
| | 1.6 | 25.90 | 26.53 | 27.01 | 26.95 | 26.93 | 27.41 | 29.07 | 25.10 | **29.32** |
| | 0.8 | 29.65 | 29.87 | 30.68 | 30.59 | 30.60 | **30.85** | 30.50 | 26.72 | 30.62 |
| **Pepper** | 1.2 | 29.06 | 29.43 | 30.03 | 29.93 | 29.95 | 30.16 | 30.39 | 26.65 | **30.48** |
| | 1.6 | 28.50 | 28.92 | 29.31 | 29.20 | 29.17 | 29.37 | 30.10 | 26.55 | **30.19** |
| | 0.8 | 21.78 | 22.03 | 23.21 | 23.07 | 23.09 | **23.59** | 23.10 | 19.82 | 23.39 |
| **PPT** | 1.2 | 21.17 | 21.56 | 22.44 | 22.30 | 22.30 | 22.65 | 23.13 | 19.76 | **23.45** |
| | 1.6 | 20.59 | 21.01 | 21.60 | 21.47 | 21.39 | 21.68 | **23.27** | 19.72 | 23.09 |
| | 0.8 | 24.49 | 24.92 | 26.47 | 26.47 | 26.44 | **27.02** | 26.18 | 21.35 | 26.93 |
| **Zebra** | 1.2 | 23.55 | 24.20 | 25.22 | 25.20 | 25.17 | 25.71 | 26.45 | 21.31 | **27.07** |
| | 1.6 | 22.71 | 23.39 | 23.99 | 23.94 | 23.88 | 24.30 | **26.70** | 21.28 | 26.57 |
| | 0.8 | 24.10 | 24.29 | 24.80 | 24.81 | 24.82 | 24.88 | 24.62 | 22.62 | **24.94** |
| **Bridge** | 1.2 | 23.65 | 23.94 | 24.33 | 24.35 | 24.38 | 24.42 | 24.77 | 22.63 | **24.99** |
| | 1.6 | 23.22 | 23.52 | 23.78 | 23.79 | 23.79 | 23.85 | **24.85** | 22.61 | 24.80 |
| | 0.8 | 25.32 | 25.54 | 26.29 | 26.33 | 26.35 | **26.60** | 26.10 | 23.45 | 26.43 |
| **Man** | 1.2 | 24.82 | 25.16 | 25.72 | 25.73 | 25.78 | 25.96 | 26.26 | 23.46 | **26.49** |
| | 1.6 | 24.36 | 24.71 | 25.09 | 25.06 | 25.07 | 25.22 | **26.36** | 23.46 | 26.27 |
| | 0.8 | 25.99 | 26.26 | 27.17 | 27.18 | 27.19 | **27.52** | 26.95 | 24.39 | 27.36 |
| **Average** | 1.2 | 25.35 | 25.77 | 26.42 | 26.39 | 26.41 | 26.68 | 27.05 | 23.82 | **27.38** |
| | 1.6 | 24.78 | 25.21 | 25.60 | 25.56 | 25.55 | 25.76 | **27.06** | 23.76 | **27.06** |

strong robustness to ranged blurring kernels.

- Upscaling factor

In **Table. V**, results concerning various upscaling factors are exhibited. Although mutual degeneration trend can be observed with upscaling factor increased from 2 to 4, our method still outperforms other state-of-the-art approaches in average with highest PSNR values, while Kim's method and Yang's method perform superior to ours on several images under conditions of upscaling factor 2 and 4 respectively. On the whole, our method performs independent to exemplars and realize state-of-the-art and even better performance for range upscaling factors, demonstrating the superior robustness to upscaling factors.

TABLE V COMPARISON OF RECONSTRUCTION PERFORMANCE WITH VIROUS $k$. THE BEST RESULTS ARE SHOWN IN BOLD.

| | $k$ | Bicubic | Sun [16] | Zeyde [31] | Timofte [18] | Yang [28] | Kim [14] | Yang [17] | Dong [36] | Proposed |
|---|---|---|---|---|---|---|---|---|---|---|
| | 2 | 32.28 | 33.17 | 33.06 | 33.06 | 33.06 | 33.21 | 35.45 | 32.21 | **35.72** |
| **Baby** | 3 | 30.91 | 31.47 | 32.12 | 32.11 | 32.11 | 32.22 | 32.86 | 28.32 | **33.22** |
| | 4 | 29.54 | 29.80 | 30.89 | 30.85 | 30.66 | 31.04 | 30.79 | 26.16 | **31.24** |
| | 2 | 30.33 | 31.31 | 31.31 | 31.25 | 31.20 | 31.49 | 34.35 | 30.38 | **34.46** |
| **Bird** | 3 | 28.72 | 29.29 | 30.06 | 30.06 | 30.00 | 30.29 | 30.89 | 26.26 | **31.19** |
| | 4 | 27.18 | 27.43 | 28.43 | 28.51 | 28.38 | 28.63 | 28.37 | 23.88 | **28.86** |
| | 2 | 22.67 | 23.68 | 23.94 | 23.73 | 23.71 | 24.10 | 28.06 | 24.13 | **28.17** |
| **Butterfly** | 3 | 21.33 | 21.95 | 22.93 | 22.86 | 22.96 | 23.76 | 23.88 | 20.23 | **24.75** |
| | 4 | 20.09 | 20.41 | 21.62 | 21.58 | 21.66 | 22.40 | 21.56 | 18.29 | **22.54** |
| | 2 | 30.15 | 30.47 | 30.52 | 30.53 | 30.55 | 30.56 | 31.34 | 30.23 | **31.57** |
| **Head** | 3 | 29.43 | 29.67 | 29.95 | 30.00 | 30.00 | 30.08 | 30.25 | 28.08 | **30.41** |
| | 4 | 28.72 | 28.85 | 29.25 | 29.29 | 29.25 | 29.37 | 29.20 | 26.68 | **29.42** |
| | 2 | 27.11 | 28.03 | 28.19 | 28.13 | 28.07 | 28.35 | **32.08** | 27.47 | 32.06 |
| **Woman** | 3 | 25.66 | 26.24 | 27.24 | 27.20 | 27.22 | 27.56 | 28.33 | 23.46 | **28.79** |
| | 4 | 24.37 | 24.64 | 25.84 | 25.80 | 25.71 | 26.24 | 25.82 | 21.53 | **26.43** |
| **Baboon** | 2 | 21.35 | 21.54 | 21.67 | 21.70 | 21.72 | 21.76 | **22.69** | 21.88 | 22.64 |
| | 3 | 20.75 | 20.89 | 21.03 | 21.08 | 21.09 | 21.13 | 21.24 | 20.37 | **21.32** |
| | 4 | 20.28 | 20.33 | 20.49 | 20.52 | 20.52 | 20.54 | 20.48 | 19.68 | **20.57** |
| **Barbara** | 2 | 24.80 | 25.10 | 25.26 | 25.23 | 25.25 | 25.33 | **26.48** | 25.12 | **26.48** |
| | 3 | 24.04 | 24.27 | 24.89 | 24.60 | 24.63 | 24.68 | 24.90 | 22.86 | **25.14** |
| | 4 | 23.33 | 23.44 | 23.79 | 23.82 | 23.80 | 23.98 | 23.79 | 21.69 | **24.00** |
| **Coastguard** | 2 | 25.53 | 25.94 | 26.32 | 26.32 | 26.40 | 26.45 | 28.47 | 26.51 | **28.61** |
| | 3 | 24.60 | 24.85 | 25.26 | 25.21 | 25.35 | 25.34 | 25.54 | 23.93 | **25.73** |
| | 4 | 23.92 | 24.03 | 24.30 | 24.32 | 24.25 | 24.38 | 24.36 | 22.99 | **24.50** |
| **Comic** | 2 | 21.84 | 22.35 | 22.60 | 22.61 | 22.62 | 22.76 | 25.13 | 22.83 | **25.21** |
| | 3 | 20.73 | 21.05 | 21.49 | 21.57 | 21.62 | 21.76 | 22.04 | 19.72 | **22.33** |
| | 4 | 19.81 | 19.94 | 20.44 | 20.47 | 20.47 | 20.56 | 20.43 | 18.27 | **20.74** |
| **Flowers** | 2 | 25.52 | 26.11 | 26.28 | 26.24 | 26.25 | 26.39 | 28.45 | 26.16 | **28.50** |
| | 3 | 24.25 | 24.62 | 25.15 | 25.18 | 25.21 | 25.35 | 25.68 | 22.96 | **25.98** |
| | 4 | 23.17 | 23.35 | 23.96 | 24.01 | 23.98 | 24.16 | 23.91 | 21.31 | **24.30** |
| **Foreman** | 2 | 29.32 | 30.25 | 30.40 | 30.24 | 30.24 | 30.48 | 32.37 | 30.32 | **33.66** |
| | 3 | 28.23 | 28.81 | 29.75 | 29.71 | 29.63 | 30.30 | 30.82 | 27.11 | **31.11** |
| | 4 | 27.22 | 27.51 | 28.74 | 28.67 | 28.52 | **29.35** | 28.06 | 25.38 | 29.15 |
| **Lena** | 2 | 29.91 | 30.48 | 30.62 | 30.60 | 30.58 | 30.75 | 32.50 | 29.73 | **32.54** |
| | 3 | 28.81 | 29.23 | 29.84 | 29.86 | 29.88 | 30.14 | 30.49 | 26.45 | **30.63** |
| | 4 | 27.78 | 27.99 | 28.76 | 28.83 | 28.70 | **29.11** | 28.81 | 24.66 | 29.02 |
| **Monarch** | 2 | 28.03 | 29.00 | 29.18 | 29.04 | 29.02 | 29.38 | 32.93 | 29.11 | **33.13** |
| | 3 | 26.63 | 27.22 | 28.07 | 28.04 | 28.10 | 28.70 | 28.95 | 25.14 | **29.75** |
| | 4 | 25.35 | 25.64 | 26.68 | 26.71 | 26.77 | 27.25 | 26.62 | 23.13 | **27.54** |
| **Pepper** | 2 | 30.02 | 30.54 | 30.65 | 30.57 | 30.58 | 30.74 | 31.49 | 29.50 | **31.76** |
| | 3 | 29.06 | 29.43 | 30.03 | 29.93 | 29.95 | 30.16 | 30.39 | 26.65 | **30.48** |
| | 4 | 28.04 | 28.23 | 29.06 | 28.99 | 28.79 | **29.25** | 29.00 | 24.75 | 29.19 |
| **PPT** | 2 | 22.50 | 23.18 | 23.61 | 23.44 | 23.47 | 23.73 | 26.79 | 23.42 | **26.81** |
| | 3 | 21.17 | 21.56 | 22.44 | 22.30 | 22.30 | 22.65 | 23.13 | 19.76 | **23.45** |
| | 4 | 19.98 | 20.14 | 21.01 | 20.86 | 20.73 | 21.24 | 21.00 | 18.20 | **21.35** |
| **Zebra** | 2 | 25.13 | 26.23 | 26.37 | 26.27 | 26.23 | 26.49 | 30.82 | 25.58 | **31.07** |
| | 3 | 23.55 | 24.20 | 25.22 | 25.20 | 25.17 | 25.71 | 26.45 | 21.31 | **27.07** |
| | 4 | 21.94 | 22.22 | 23.43 | 23.48 | 23.43 | 23.98 | 23.33 | 19.27 | **24.06** |
| **Bridge** | 2 | 24.54 | 24.97 | 25.18 | 25.17 | 25.20 | 25.27 | 27.12 | 25.18 | **27.13** |
| | 3 | 23.65 | 23.94 | 24.33 | 24.35 | 24.38 | 24.42 | 24.77 | 22.63 | **24.99** |
| | 4 | 22.80 | 22.93 | 23.39 | 23.39 | 23.37 | 23.38 | 23.35 | 21.39 | **23.60** |
| **Man** | 2 | 25.80 | 26.29 | 26.51 | 26.50 | 26.49 | 26.64 | **28.60** | 26.36 | 28.59 |
| | 3 | 24.82 | 25.16 | 25.72 | 25.73 | 25.78 | 25.96 | 26.26 | 23.46 | **26.49** |
| | 4 | 23.95 | 24.11 | 24.79 | 24.78 | 24.77 | 25.00 | 24.73 | 21.93 | **25.06** |
| **Average** | 2 | 26.49 | 27.15 | 27.32 | 27.26 | 27.26 | 27.44 | 29.73 | 27.01 | **29.90** |
| | 3 | 25.35 | 25.77 | 26.42 | 26.39 | 26.41 | 26.68 | 27.05 | 23.82 | **27.38** |
| | 4 | 24.308 | 24.50 | 25.27 | 25.27 | 25.21 | 25.55 | 25.20 | 22.18 | **25.64** |

● Noise intensity

**Table.** VI shows the reconstruction performance with various noise intensities. As analyzed in Section II, the patch-wise slope-based implementation of our method performs innate suppression of noise, which can be demonstrated by the superior performance with highest PSNR in average. Although sparsity-deduced approaches like Kim's and Yang's methods have advantages in noise suppression, our method still outperforms them except for several conditions where our method performs slightly inferior.

TABLE VI COMPARISON OF RECONSTRUCTION PERFORMANCE WITH VIROUS NOISE INTENSITIES. THE BEST RESULTS ARE SHOWN IN BOLD.

| | $\sigma_N$ | Bicubic | Sun [16] | Zeyde [31] | Timofte [18] | Yang [28] | Kim [14] | Yang [17] | Dong [36] | Proposed |
|---|---|---|---|---|---|---|---|---|---|---|
| **Baby** | 0 | 30.91 | 31.47 | 32.12 | 32.11 | 32.11 | 32.22 | 32.86 | 28.32 | **33.22** |
| | 2 | 30.78 | 31.32 | 31.93 | 31.91 | 31.89 | 32.02 | 32.57 | 28.23 | **32.76** |
| | 4 | 30.43 | 30.90 | 31.42 | 31.37 | 31.30 | 31.48 | **31.87** | 27.94 | 31.71 |
| **Bird** | 0 | 28.72 | 29.29 | 30.06 | 30.06 | 30.00 | 30.29 | 30.89 | 26.26 | **31.19** |
| | 2 | 28.64 | 29.19 | 29.91 | 29.91 | 29.84 | 30.14 | 30.67 | 26.18 | **30.88** |
| | 4 | 28.43 | 28.93 | 29.61 | 29.57 | 29.47 | 29.77 | 30.14 | 25.98 | **30.15** |
| **Butterfly** | 0 | 21.33 | 21.95 | 22.93 | 22.86 | 22.96 | 23.76 | 23.88 | 20.23 | **24.75** |
| | 2 | 21.32 | 21.93 | 22.91 | 22.83 | 22.94 | 23.72 | 23.83 | 20.19 | **24.68** |
| | 4 | 21.28 | 21.88 | 22.85 | 22.76 | 22.86 | 23.62 | 23.73 | 20.14 | **24.49** |
| **Head** | 0 | 29.43 | 29.67 | 29.95 | 30.00 | 30.00 | 30.08 | 30.25 | 28.08 | **30.41** |
| | 2 | 29.33 | 29.56 | 29.83 | 29.86 | 29.85 | 29.96 | 30.07 | 27.91 | **30.13** |
| | 4 | 29.07 | 29.27 | 29.47 | 29.48 | 29.44 | **29.59** | **29.59** | 27.55 | 29.44 |
| **Woman** | 0 | 25.66 | 26.24 | 27.24 | 27.20 | 27.22 | 27.56 | 28.33 | 23.46 | **28.79** |
| | 2 | 25.63 | 26.20 | 27.19 | 27.14 | 27.16 | 27.50 | 28.22 | 23.40 | **28.62** |
| | 4 | 25.51 | 26.05 | 26.99 | 26.95 | 26.93 | 27.27 | 27.90 | 23.30 | **28.14** |
| **Baboon** | 0 | 20.75 | 20.89 | 21.03 | 21.08 | 21.09 | 21.13 | 21.24 | 20.37 | **21.32** |
| | 2 | 20.74 | 20.87 | 21.01 | 21.06 | 21.08 | 21.11 | 21.22 | 20.34 | **21.29** |
| | 4 | 20.71 | 20.83 | 20.97 | 21.01 | 21.02 | 21.06 | 21.15 | 20.29 | **21.20** |
| **Barbara** | 0 | 24.04 | 24.27 | 24.89 | 24.60 | 24.63 | 24.68 | 24.90 | 22.86 | **25.14** |
| | 2 | 24.01 | 24.24 | 24.55 | 24.56 | 24.59 | 24.65 | 24.85 | 22.83 | **25.07** |
| | 4 | 23.94 | 24.16 | 24.45 | 24.46 | 24.48 | 24.54 | 24.71 | 22.73 | **24.85** |
| **Coastguard** | 0 | 24.60 | 24.85 | 25.26 | 25.21 | 25.35 | 25.34 | 25.54 | 23.93 | **25.73** |
| | 2 | 24.58 | 24.83 | 25.22 | 25.17 | 25.31 | 25.30 | 25.49 | 23.88 | **25.63** |
| | 4 | 24.48 | 24.72 | 25.09 | 25.03 | 25.15 | 25.17 | 25.31 | 23.78 | **25.38** |
| **Comic** | 0 | 20.73 | 21.05 | 21.49 | 21.57 | 21.62 | 21.76 | 22.04 | 19.72 | **22.33** |
| | 2 | 20.72 | 21.03 | 21.47 | 21.54 | 21.60 | 21.73 | 22.01 | 19.70 | **22.29** |
| | 4 | 20.68 | 20.99 | 21.42 | 21.49 | 21.54 | 21.67 | 21.92 | 19.67 | **22.18** |
| **Flowers** | 0 | 24.25 | 24.62 | 25.15 | 25.18 | 25.21 | 25.35 | 25.68 | 22.96 | **25.98** |
| | 2 | 24.22 | 24.59 | 25.11 | 25.13 | 25.17 | 25.31 | 25.62 | 22.93 | **25.89** |
| | 4 | 24.14 | 24.49 | 25.00 | 25.01 | 25.03 | 25.18 | 25.45 | 22.83 | **25.64** |
| **Foreman** | 0 | 28.23 | 28.81 | 29.75 | 29.71 | 29.63 | 30.30 | 30.82 | 27.11 | **31.11** |
| | 2 | 28.17 | 28.73 | 29.65 | 29.60 | 29.52 | 30.17 | 30.62 | 27.03 | **30.81** |
| | 4 | 27.99 | 28.51 | 29.35 | 29.29 | 29.17 | 29.81 | **30.12** | 26.81 | 30.07 |
| **Lena** | 0 | 28.81 | 29.23 | 29.84 | 29.86 | 29.88 | 30.14 | 30.49 | 26.45 | **30.63** |
| | 2 | 28.73 | 29.14 | 29.72 | 29.73 | 29.75 | 30.01 | 30.32 | 26.36 | **30.36** |
| | 4 | 28.50 | 28.87 | 29.39 | 29.38 | 29.36 | 29.65 | **29.86** | 26.16 | 29.68 |
| **Monarch** | 0 | 26.63 | 27.22 | 28.07 | 28.04 | 28.10 | 28.70 | 28.95 | 25.14 | **29.75** |
| | 2 | 26.58 | 27.16 | 27.98 | 27.95 | 28.00 | 28.61 | 28.84 | 25.12 | **29.53** |
| | 4 | 26.44 | 26.99 | 27.76 | 27.72 | 27.74 | 28.35 | 28.52 | 24.96 | **28.96** |
| **Pepper** | 0 | 29.06 | 29.43 | 30.03 | 29.93 | 29.95 | 30.16 | 30.39 | 26.65 | **30.48** |
| | 2 | 28.98 | 29.33 | 29.92 | 29.81 | 29.81 | 30.04 | 30.23 | 26.57 | 30.22 |
| | 4 | 28.73 | 29.04 | 29.56 | 29.44 | 29.40 | 29.67 | **29.76** | 26.35 | 29.53 |
| **PPT** | 0 | 21.17 | 21.56 | 22.44 | 22.30 | 22.30 | 22.65 | 23.13 | 19.76 | **23.45** |
| | 2 | 21.16 | 21.56 | 22.43 | 22.29 | 22.29 | 22.65 | 23.11 | 19.75 | **23.41** |
| | 4 | 21.13 | 21.53 | 22.39 | 22.25 | 22.24 | 22.61 | 23.06 | 19.73 | **23.33** |
| **Zebra** | 0 | 23.55 | 24.20 | 25.22 | 25.20 | 25.17 | 25.71 | 26.45 | 21.31 | **27.07** |
| | 2 | 23.52 | 24.18 | 25.18 | 25.15 | 25.13 | 25.66 | 26.37 | 21.29 | **26.95** |
| | 4 | 23.45 | 24.09 | 25.05 | 25.03 | 24.99 | 25.50 | 26.15 | 21.22 | **26.62** |
| **Bridge** | 0 | 23.65 | 23.94 | 24.33 | 24.35 | 24.38 | 24.42 | 24.77 | 22.63 | **24.99** |
| | 2 | 23.62 | 23.91 | 24.29 | 24.31 | 24.34 | 24.40 | 24.71 | 22.61 | **24.92** |
| | 4 | 23.55 | 23.83 | 24.19 | 24.20 | 24.22 | 24.29 | 24.56 | 22.52 | **24.71** |
| **Man** | 0 | 24.82 | 25.16 | 25.72 | 25.73 | 25.78 | 25.96 | 26.26 | 23.46 | **26.49** |
| | 2 | 24.78 | 25.12 | 25.66 | 25.66 | 25.71 | 25.92 | 26.18 | 23.43 | **26.38** |
| | 4 | 24.70 | 25.01 | 25.54 | 25.53 | 25.56 | 25.77 | 25.99 | 23.30 | **26.11** |
| **Average** | 0 | 25.35 | 25.77 | 26.42 | 26.39 | 26.41 | 26.68 | 27.05 | 23.82 | **27.38** |
| | 2 | 25.31 | 25.72 | 26.33 | 26.31 | 26.33 | 26.61 | 26.94 | 23.76 | **27.21** |
| | 4 | 25.18 | 25.56 | 26.14 | 26.11 | 26.11 | 26.39 | 26.66 | 23.63 | **26.79** |

## V. CONCLUSIONS AND FUTURE WORK

In this paper, we propose a fast and simple single image super-resolution algorithm based on sigmoid transformation, utilize the overlapped patch-wise sigmoid transformation to realize slope-based sharpening of the image, and implement the sharpening operation as an imposed sharpening regularization term in the reconstruction. Extensive experiments compared with other state-of-the-art approaches demonstrate the superiority of the proposed algorithm in effectiveness and efficiency. Considering the fast, simple and effective SR processing of the proposed method with no need for learning or training, it has widespread application value and prospect.

Although the proposed overlapped patch-wise sigmoid transformation realizes state-of-the-art and even better reconstruction performance, the parameters utilized including steepest parameter $K$ and location parameter $B$ are empirically determined, which can be further improved. In the future, we will investigate the adaptive determination of the parameters to enhance the adaptability of sigmoid transformation to complex images.